\documentclass{article}

\usepackage[T1]{fontenc}
\usepackage[utf8]{inputenc}
\usepackage{lmodern}

\usepackage{amsmath,amssymb,amsfonts}

\usepackage{graphicx}
\usepackage{booktabs}
\usepackage{tabularx}
\usepackage{array}
\usepackage{mathtools}
\usepackage{makecell}
\usepackage{float}
\usepackage{xcolor}
\usepackage[final]{microtype}
\usepackage{xstring}
\usepackage{placeins}
\usepackage[ruled,vlined]{algorithm2e}
\SetKw{KwState}{State}
\DontPrintSemicolon
\usepackage[
  letterpaper,
  left=1.25in,
  right=1.25in,
  top=1.2in,
  bottom=1.2in
]{geometry}

\usepackage{natbib}
\usepackage[colorlinks=true,allcolors=blue]{hyperref}

\makeatletter
\newcommand{\myparagraph}[1]{%
  \paragraph{\@addpunct{#1}}%
}
\makeatother

\title{When F$_1$ Fails: Granularity-Aware Evaluation for Dialogue Topic Segmentation}

\author{
Michael Coen\\
Independent Researcher\\
\texttt{mhcoen@alum.mit.edu}
}

\date{Preprint. Under review.}

\begin{document}
\maketitle

\begin{abstract}
Dialogue topic segmentation supports summarization, retrieval, memory
management, and conversational continuity. Despite decades of prior
work, evaluation practice remains framed as boundary prediction,
using strict matching and F1-based metrics at a fixed threshold. This
obscures boundary density and granularity effects—factors that matter
as LLM-based conversational systems increasingly rely on segmentation
to manage history beyond fixed context windows. Without structure,
accumulated dialogue degrades efficiency and coherence.

This paper introduces an evaluation perspective for dialogue topic
segmentation that reports boundary density and segment alignment
diagnostics (purity and coverage) alongside window-tolerant F1
(W-F1). By explicitly analyzing boundary density through separating
boundary scoring (per-position scores) from boundary selection
(mapping scores to a boundary set), we evaluate segmentation quality
across density regimes rather than at a single threshold
setting. Through cross-dataset empirical evaluation, we show that
reported performance differences across dialogue segmentation
benchmarks are often not driven by boundary placement quality alone;
under common evaluation protocols, annotation granularity mismatch and
sparse boundary labels can dominate these differences.

We evaluate multiple segmentation strategies across eight dialogue
datasets spanning task-oriented, open-domain, meeting-style, and
synthetic interactions. Across these settings, boundary-based metrics
are strongly coupled to boundary density: threshold sweeps produce
larger W-F1 changes than switching between methods. These findings
support viewing topic segmentation as a granularity selection problem
and motivate treating boundary selection as a distinct, tunable step.
\end{abstract}

\section{Introduction}
Dialogue topic segmentation is commonly framed as identifying points
in a conversation where ``the subject changes.'' This framing is
appealingly simple. It is also incorrect. Topic change is often
gradual, topic structure is often nested, and both depend on
perspective.

Despite this, many dialogue segmentation benchmarks treat annotated
boundaries as objective ground truth and evaluate systems using strict
boundary matching and F1-based metrics~\cite{vanrijsbergen1979}. These
metrics collapse distinct failure modes into a single number. Small
boundary shifts are often harmless when the goal is coherent topic
blocks, while missing or adding a few boundaries can be catastrophic
if boundaries drive downstream actions such as summarizing or dropping
earlier turns. As we show in this paper, these effects are tightly coupled to
segmentation granularity and boundary density, which are left implicit
by standard boundary-based metrics.

A central difficulty is that segmentation granularity is latent in
standard evaluation. Most benchmarks compare systems at a single
operating point (i.e., a set of fixed thresholds or selection
settings), even though the number of boundaries produced varies across
methods and datasets without explicit control. As a result,
boundary-based metrics often reflect alignment in boundary
\emph{density} rather than boundary placement quality. To make
granularity explicit, this work separates boundary scoring from
boundary selection and evaluates performance across boundary densities
rather than at a single threshold.

This paper addresses that mismatch by reframing what gets reported and
what gets optimized. The evaluation objective in this paper jointly
reports (i) window-tolerant F1 (W-F1), (ii) boundary density, and
(iii) segment alignment diagnostics (purity and coverage). An
accompanying algorithmic contribution separates boundary scoring from
boundary selection, making boundary density explicit and controllable.
Together, these contributions distinguish detection failures from
granularity mismatches (i.e., discrepancies between the segmentation
resolution at which boundaries are produced and that implied by the
annotation scheme) and explain why a single global threshold fails to
transfer across datasets.

We distinguish between two notions that are often conflated in
boundary-based evaluation. \emph{Boundary placement alignment} refers
to whether predicted boundaries occur at the same locations as gold
boundaries (within a tolerance window), which is the intended target
of metrics such as F1 and W-F1. By contrast, \emph{boundary density
alignment} refers only to whether the number of predicted boundaries
matches the number implied by the annotation scheme. We refer to this
implicit granularity level as the annotation \emph{regime} (e.g.,
coarse thematic divisions versus finer chapter- or subtopic-level
segmentation), independent of which structural segmentation markers
(e.g., chapters, sections, turns) are used to realize those boundaries.
\emph{Segmentation granularity} is a modeling- and selection-level
property, referring to the resolution at which topic boundaries are
produced. It determines boundary density through the boundary selection
rule, sometimes implicitly, and thereby strongly influences
boundary-based metrics.

A single-number boundary F1 can conflate two distinct factors: (i)
localization (whether predicted boundaries fall near annotated topic
transitions), and (ii) segmentation granularity (how many boundaries
are produced overall).  When the granularity of predicted segmentations
differs from that implied by the annotation scheme, apparent F1 gains
can arise primarily from matching the number of boundaries rather than
from improved boundary placement at a matched resolution. To make this
distinction explicit, we report complementary diagnostics that
separately quantify boundary localization, boundary density alignment,
and gold-relative segment alignment alongside tolerant boundary
accuracy scores; formal definitions are given in
Section~\ref{sec:evaluation-metrics}.

Standard boundary-based evaluation implicitly assumes that, once
boundary density is controlled, differences in F1 or W-F1 primarily
reflect differences in boundary placement quality rather than
thresholding or selection behavior.  A central empirical finding of
this work is that \emph{boundary-based evaluation metrics can be dominated
by segmentation granularity induced by boundary selection, rather than
by boundary detection quality.}  In particular, for a fixed scoring
model, sweeping the boundary selection threshold produces larger
changes in W-F1 than switching between structurally distinct
segmentation methods, a pattern made explicit by density--quality
curves in Section~\ref{sec:results}
(Figure~\ref{fig:density-quality}).  As a result, leaderboard-style
comparisons that report only a single operating point can conflate
method improvements with shifts in boundary density.

The proposed separation of boundary scoring and boundary
selection motivates alternative ways of reasoning about the
construction, tuning, and evaluation of segmentation models under
varying annotation granularities.

\begin{enumerate}
\item We show that boundary-based metrics (including F1 and W-F1) can
  be dominated by boundary density alignment: even non-semantic
  baselines achieve competitive scores when their output density
  matches the gold annotation.
\item We propose a composite evaluation objective that reports
  boundary density and purity/coverage alongside W-F1, exposing
  failure modes hidden by standard boundary metrics.
\item We demonstrate, via density--quality curves, that sweeping
  boundary selection thresholds for a fixed scoring model produces
  substantially larger changes in W-F1 and coverage than switching
  between segmentation methods, making granularity mismatch explicit.
\item We show that a separation between boundary scoring and boundary
  selection is necessary to treat boundary density as a controllable
  variable rather than an incidental consequence of threshold tuning.
\end{enumerate}

\subsection{Motivation: Conversational Memory in AI Chat Systems}

Large language model (LLM) based chat systems operate under finite context
windows, which require practical systems to manage conversational history
explicitly. As conversations grow longer, earlier turns must be dropped,
summarized, or selectively retrieved. Topic segmentation decisions therefore
directly determine which portions of prior dialogue are preserved, compressed,
or reintroduced.

Importantly, this challenge does not disappear as context windows grow larger.
Unstructured accumulation of earlier dialogue increases computational cost and
dilutes relevance, reducing the model's ability to focus on salient information.
Effective conversational systems therefore require explicit structure over
dialogue history rather than unbounded context expansion~\cite{liu-etal-2024-lost}.

In this setting, topic segmentation functions as a memory control
mechanism rather than a purely linguistic task. Poor segmentation can
cause relevant context to be discarded prematurely or irrelevant
context to be retained, leading to degraded long-horizon coherence and
unstable behavior across turns, e.g.,~\citep{sclar-etal-2024-quantifying,
  razavi-etal-2025-promptset}.  The lack of robust long-horizon
memory remains an open challenge for current LLM-based systems,
particularly in conversations spanning multiple sessions, motivating
careful attention to segmentation decisions and their evaluation.

This work is motivated by the development of \emph{Episodic}, a
conversational memory system that organizes dialogue into
topic-structured representations to support persistent interactions
with LLM-based assistants~\cite{episodic_github}. Within such
systems, boundary placement governs summarization, context selection,
and context reintegration, making segmentation granularity a
first-order systems design decision rather than an evaluation detail.
In Episodic, topic segmentation interacts with neural modules
for context selection, compression of accumulated history, response
generation, critique, and synthesis.

\myparagraph{Motivating observation from Episodic}
In early prototyping of Episodic, modest changes to
boundary-selection parameters (and thus boundary density) produced
immediately visible changes in the context assembled for the LLM and
in the resulting assistant responses, even when tolerant boundary F1
changed little. Increasing density produced shorter context chunks
used for LLM conditioning, which truncated cross-turn dependencies
and yielded incomplete summaries. Decreasing density, by contrast,
produced coarser context units that reduced resolution around topic
transitions and, in some cases, suppressed boundary evidence to the
point that topic transitions were not detected at all.

This effect is expected because segmentation directly determines the
units selected for context reintegration (and, when needed,
summarization), thereby changing the conditioning input presented
within the finite context window. This effect was compounded in
multi-session settings. Maintaining topic continuity required
concatenating multiple same-topic segments drawn from temporally
distinct conversations. Higher density amplified fragmentation, while
lower density increased topic mixing.  These observations motivated
treating boundary density as an explicit control variable and
reporting purity/coverage to distinguish within-gold refinement from
cross-gold mixing and fragmentation.\footnote{We do not report a
separate downstream evaluation here; this paper focuses on the
evaluation and diagnostic methodology motivated by these
observations.}

\section{Background and Related Work}

Topic segmentation has a long history in text processing, with
evaluation practice, dataset design, and granularity assumptions
evolving somewhat independently. We organize prior work along these
three dimensions.

\subsection{Boundary-Based Evaluation Metrics}
Topic segmentation evaluation has historically centered on boundary
alignment. Early metrics such as $P_k$~\cite{beeferman1999statistical} penalize
segmentations where predicted and reference boundaries fall in different
segments under a sliding window. WindowDiff~\cite{pevzner2002critique} refines
this approach to address known biases in $P_k$. Both metrics are typically
computed against a single reference segmentation and measure deviation from
that reference, even though multiple segmentations (and granularities) can be
defensible~\citep{fournier-inkpen-2012-segmentation,fournier-2013-evaluating}.
More recent dialogue segmentation work has largely adopted precision, recall,
and exact-match $F_1$ computed over boundary positions \citep{jiang2023superdialseg,xia2022dialogue}.
These boundary-only scores likewise treat the available annotation as the
evaluation target; when annotation granularity varies across datasets or differs
from system behavior, they can yield misleading comparisons.

\myparagraph{Beyond window-based and exact-match boundary scores}
A substantial line of work in \emph{text segmentation evaluation} argues that
(1) strict boundary matching over-penalizes near misses, and
(2) window-based metrics such as $P_k$ and WindowDiff introduce task- and
parameter-dependent biases, motivating edit-distance and confusion-matrix-based
alternatives. \citet{fournier-inkpen-2012-segmentation} propose
\emph{segmentation similarity} ($S$), an edit-distance--based similarity that is
symmetric (does not privilege a single ``true'' reference) and supports
evaluation against multiply-coded corpora. \citet{fournier-2013-evaluating}
extends this direction with \emph{boundary edit distance} and
\emph{boundary similarity} ($B$), yielding a principled way to obtain
boundary-level IR-style quantities (precision/recall/$F$) while awarding graded
credit for near misses. Relatedly, \citet{scaiano-inkpen-2012-getting} derive a
boundary confusion matrix from WindowDiff-style windows (WinPR), separating
false positives from false negatives while retaining tunable near-miss
sensitivity. For corpora with multiple annotators or where segmentation is
treated as a \emph{unitizing} task, agreement measures that jointly align and
score units (e.g., $\gamma$) provide an alternative to single-reference
evaluation~\citep{mathet2015gamma}.

\subsection{Dialogue Segmentation Datasets}
\label{sec:relateddatasets}
Dialogue segmentation benchmarks differ not only in domain but in what counts as a
\emph{topic boundary}: some datasets provide explicit segment boundaries, others provide
weaker supervisory signals (e.g., section/domain labels) that must be repurposed, and
some evaluation sets are synthetically constructed via concatenation.
These design choices induce large, systematic differences in boundary density and implied
granularity, so boundary-only metrics can confound localization with density alignment.
Section~\ref{sec:datasets} summarizes the eight datasets used in our experiments and the
exact gold-boundary definitions used after canonicalization.

\subsection{Segmentation Granularity}
Classic segmentation methods such as TextTiling~\cite{hearst1997texttiling} and
Choi-style benchmarks~\cite{choi2000advances} established the paradigm of
predicting a single boundary set evaluated against gold annotations. More
recent work has also reframed segmentation as supervised boundary prediction on
large automatically-derived corpora (e.g., Wikipedia section boundaries),
typically evaluated with boundary- and window-based metrics
\citep{koshorek2018text}. Across these settings, research has refined scoring
models and architectures, but the assumption that a single ``correct''
granularity exists has remained largely unexamined. Prior work rarely treats
segmentation granularity as a first-class, controllable variable: boundary
density is typically a side effect of threshold tuning rather than an explicit
design choice, and evaluation protocols do not distinguish between detection
failures (placing boundaries at incorrect locations) and granularity mismatches
(producing a different number of boundaries than the annotation scheme implies).
This paper addresses that gap by separating boundary scoring from boundary
selection and proposing evaluation criteria that make granularity effects
explicit.

\myparagraph{Multiple valid granularities and weakly-identified boundaries}
Empirical studies of human segmentation consistently emphasize that topical
structure is only weakly identified by surface form: topic change is gradual,
the task is under-defined, and annotators operate at different granularities.
In a large-scale topical segmentation study, \citet{kazantseva-szpakowicz-2012-topical}
report low overall inter-annotator agreement but substantially higher agreement
on a subset of \emph{prominent} breaks, motivating prominence-weighted
evaluation. More broadly, discourse segmentation work notes weak consensus on
what the relevant units and criteria should be, even when humans can be
reliable under a fixed criterion~\citep{passonneau-litman-1997-discourse}. These
findings support the premise of this paper: benchmark boundary sets reflect one
operational granularity among many, so boundary-only scores can conflate
boundary \emph{localization} with \emph{granularity/density alignment}.

\section{Problem Setting}
\label{sec:problem-setting}
Topic boundaries are not intrinsic properties of conversations; they
are defined relative to a chosen segmentation granularity, which
reflects task requirements and user or annotator preferences about
where topical divisions should be made. A model that produces
fine-grained discourse-phase boundaries can be operationally
reasonable yet score poorly against a dataset that annotates only
coarse task transitions. Conversely, a model tuned to sparse labels
may suppress meaningful subtopic structure in order to match the
annotated boundary density.

In contrast, this paper treats topic segmentation as two separate operations:
\begin{enumerate}
\item \textbf{Boundary scoring:} assign a score to each candidate
  boundary position.
\item \textbf{Boundary selection:} choose which candidates become
  output boundaries, controlling boundary density and spacing.
\end{enumerate}

Collapsing these operations makes boundary density an accidental
byproduct of threshold tuning. Our baseline experiments confirm that
this produces systematically misleading results: the effect is
demonstrated in
Section~\ref{sec:baselines}. Figure~\ref{fig:worked_example}
(\S~\ref{sec:worked_example}) provides a concrete worked example
of this granularity mismatch.

\section{Datasets}
\label{sec:datasets}

We evaluate across eight dialogue datasets spanning diverse annotation
regimes (see Section~\ref{sec:relateddatasets} for background on
annotation philosophies).  The purpose of cross-dataset evaluation is
not only to compare performance, but to expose how annotation
philosophy and boundary density interact with evaluation metrics. We
summarize the supervision signal used for each dataset and how it is
converted to turn-adjacent gold boundaries in
Table~\ref{tab:segmentation-philosophy} (full canonicalization details
in Appendix~D).

\myparagraph{Explicit segmentation annotation}
DialSeg711~\cite{xu2021topicaware} is constructed by joining single-topic
dialogues (from MultiWOZ and the Stanford Dialogue Dataset), so boundaries
primarily reflect splice points rather than naturally occurring gradual
topic transitions.
SuperDialseg~\cite{jiang2023superdialseg} provides large-scale supervised
segmentation built from document-grounded dialogues, with segment IDs aligned
to source document structure.
TIAGE~\cite{xie2021tiage} augments PersonaChat with per-turn topic-shift
annotations, yielding sparse shift points rather than hierarchical segment
structure.
QMSum~\cite{zhong2021qmsum} provides topic annotations for meeting transcripts;
because QMSum allows topic-relevant spans to be non-contiguous, we convert its
annotations to turn-adjacent boundary positions as described in Appendix~D.

\myparagraph{Repurposed labels}
MultiWOZ~\cite{budzianowski2018multiwoz} provides domain and dialogue-state
annotations; we derive boundaries from domain label changes.
DailyDialog~\cite{li2017dailydialog},
Taskmaster~\cite{byrne2019taskmaster}, and
Topical-Chat~\cite{gopalakrishnan2019topicalchat} do not provide explicit
segmentation boundaries; we derive boundaries from changes in available
topic/domain/task labels between consecutive utterances (Appendix~D).

Together, these datasets span sparse, medium, and dense annotation regimes,
providing a testbed for evaluating segmentation behavior under annotation
granularity shift. We refer to SuperDialseg as \emph{SuperSeg} throughout.

\begin{table}[t]
  \centering
  \small
  \setlength{\tabcolsep}{5pt}
  \begin{tabular}{lll}
    \toprule
    \textbf{Dataset} & \textbf{Supervision signal} & \textbf{Gold boundary derivation} \\
    \midrule
    \multicolumn{3}{l}{\textit{Explicit segmentation / shift labels}} \\
    \quad DialSeg711 & synthetic splice construction & splice point (segment ID change) \\
    \quad SuperSeg & document section IDs & section change \\
    \quad TIAGE & topic-shift annotations & annotated shift / segment change \\
    \quad QMSum & topic spans & boundary at adjacent span break \\
    \midrule
    \multicolumn{3}{l}{\textit{Repurposed labels (boundary = label change)}} \\
    \quad MultiWOZ & domain label & domain change \\
    \quad DailyDialog & dialogue topic label & topic change \\
    \quad Taskmaster & task/subtask labels & task/subtask change \\
    \quad Topical-Chat & conversation topic label & topic change \\
    \bottomrule
  \end{tabular}
  \caption{Supervision signals and the resulting turn-adjacent gold boundaries used in evaluation.
    Datasets are cited in the main text (§\ref{sec:datasets}); full extraction/canonicalization is in Appendix~D.}
  \label{tab:segmentation-philosophy}
\end{table}

\section{Methodology}

This section operationalizes the separation between boundary scoring
and boundary selection introduced in
Section~\ref{sec:problem-setting} for empirical evaluation.

\myparagraph{Experimental scope} The experiments reported in
Sections~\ref{sec:evaluation-metrics} and~\ref{sec:results}
intentionally evaluate only the neural boundary scoring component
under static threshold\(+\)gap selection. Episodic, as deployed, is a
hybrid system in which neural boundary scores are complemented by
additional online signals (e.g., embedding-drift–based change
detection).
These auxiliary signals are not used in the main experiments and are
held out deliberately to isolate the behavior of boundary-based
evaluation metrics under controlled changes in boundary
density. Accordingly, the performance tables and density--quality
curves should be interpreted as diagnostics of scoring and selection
behavior under controlled boundary densities.

\subsection{Topic Segmentation Algorithm}

Given a dialogue consisting of messages $m_1,\dots,m_T$, we define
candidate boundary positions $i \in \{1,\dots,T-1\}$ between adjacent
messages.

\myparagraph{Selection rule family}
Boundary selection operates over candidate positions after scoring,
and is independent of the specific scoring model used. All boundary
selection methods used in this work instantiate a common selection
primitive: a boundary at position $i$ is committed iff (i) its
accumulated evidence $e_i(t)$ exceeds a threshold $\tau(t)$, and (ii)
it is separated by at least $g$ messages from any previously committed
boundary. The \emph{static} selection rule used in the main experiments
is the special case where $e_i(t) = s_i$ for all $t$ at which position
$i$ is evaluated (no temporal accumulation) and $\tau(t)=\tau$ is
fixed. The \emph{adaptive} selection rule illustrated in
Figure~\ref{fig:adaptive_commitment} and discussed in
Section~\ref{sec:boundary-selection} generalizes this primitive by
defining (a) an online evidence accumulator $e_i(t)$ and (b) a
time-varying threshold $\tau(t)$, adjusted to enforce a target
boundary rate relative to the number of candidate positions processed
(see Appendix~F for a complete formal specification).

\myparagraph{Selection hyperparameters and transparency}
Boundary density is controlled entirely by the boundary selection rule
through the threshold $\tau$ and minimum spacing parameter $g$. For all
single-operating-point results reported in this paper
(Tables~\ref{tab:baselines}, \ref{tab:external_methods}, \ref{tab:stage3},
and~\ref{tab:purity_coverage}), $g$ is fixed at 3 and $\tau$ is fixed
at 0.50 after temperature calibration. Density--quality curves
(Section~\ref{sec:granularity}) sweep $\tau$ while holding $g$ fixed;
the adaptive controller (Section~\ref{sec:boundary-selection},
Figure~\ref{fig:adaptive_commitment}) is a separate control experiment. Neither parameter is tuned per dataset, per
split, or on held-out validation data to match gold boundary counts or
to optimize any evaluation metric. As a result, variation in boundary
density across datasets reflects differences in annotation granularity
rather than dataset-specific hyperparameter adjustment. This choice is
intentional: per-dataset tuning of $g$ or $\tau$ would partially
obscure the granularity effects this work aims to expose.

\myparagraph{Boundary scoring}
For each candidate position $i$, the scorer computes a real-valued
score $s_i \in \mathbb{R}$, where larger values indicate stronger
evidence of a topic boundary. Scores are computed using a
window-vs-window comparison: a window of turns preceding position $i$
is compared against a window of turns following $i$, rather than
relying on the adjacent message pair alone.

\myparagraph{Boundary selection}
The scorer ranks candidates; boundary selection determines which
become final boundaries. In online settings (i.e., incremental,
decision-time selection as dialogue unfolds), selection may accumulate
evidence over time before committing to a boundary. This separation
decouples final segmentation granularity from the scorer's behavior.

\myparagraph{Static boundary selection rule (used in experiments)}
\label{sec:selection-rule}
In our experiments, boundary selection reduces to a static selection
rule based on thresholding with minimum spacing.
\begin{enumerate}
\item Select all candidate positions $i$ such that $s_i \ge \tau$.
\item Enforce a minimum spacing $g$ via greedy non-maximum
  suppression: candidates are processed in descending score order, and
  a candidate i is accepted only if $|i - b| \geq g$ for all
  previously accepted boundaries $b$.
\end{enumerate}
The resulting set of topic boundaries is $B \subseteq
\{1,\dots,T-1\}$.

\noindent\textbf{Scope.}
For Figure~\ref{fig:adaptive_commitment}, we allow the threshold $\tau$
to vary over time (logged as \texttt{min\_evidence}, the running
evidence threshold used by the selection rule); all other aspects of
the selection rule remain unchanged. During evidence accumulation, each candidate's fixed score $s_i$
(computed once at first availability using frozen reference windows)
is repeatedly added to its evidence total, implementing a
time-to-threshold integrator: candidates with stronger initial
boundary signals reach the commitment threshold faster.
Freezing the score at the point of first availability ensures that
the commitment decision reflects the strength of the boundary signal
at the moment of topic departure, rather than being confounded by
subsequent context that may diverge further or revisit the original topic.

\myparagraph{Threshold sweeps for density--quality curves}
To construct density--quality curves
(Section~\ref{sec:granularity}), we sweep the selection threshold $\tau$
over a fixed linear grid while holding the spacing parameter $g$
constant at $g = 3$. Specifically, $\tau$ is swept linearly over the
interval $[0.05, 0.95]$ in increments of $0.05$, yielding 19 operating
points per method. At each value of $\tau$, boundary selection is
performed using the same static selection rule, and boundary density,
W-F1, purity, and coverage are recomputed. No other parameters are
varied during the sweep.

\myparagraph{Selection hyperparameters (reported values)}
All single-operating-point results for the proposed neural scorer
(Tables~\ref{tab:stage3} and~\ref{tab:purity_coverage}) use a fixed minimum
spacing of $g = 3$ messages and a fixed global threshold of $\tau =
0.50$ applied to temperature-calibrated scores. These values are held
constant across all datasets, splits, and runs.
Baselines in Table~\ref{tab:baselines} that directly output boundaries
(e.g., No-boundary, Periodic, Oracle-density Random) do not use $\tau$/$g$;
external methods in Table~\ref{tab:external_methods} (TextTiling, CSM)
use their own internal scoring and thresholding mechanisms evaluated at
fixed operating points (see Appendix~H for details).
Density--quality
curves (Section~\ref{sec:granularity},
Figure~\ref{fig:density-quality}) sweep $\tau$ over $[0.05, 0.95]$
while holding $g = 3$ fixed, as described in ``Threshold sweeps for
density--quality curves'' above.

\subsection{Scoring Model Architecture}
The boundary scoring model is a fine-tuned DistilBERT-based binary
classifier. For each candidate boundary position $i$, the model
encodes a left context window and a right context window around $i$
using the input format \texttt{[CLS] pre-window [SEP] post-window
  [SEP]}. Unless otherwise specified, the scoring model uses a
symmetric local context window of $k{=}4$ utterances before and after
each candidate boundary (total window size 9), truncated at dialogue
boundaries when insufficient context is available. The final hidden state
of the \texttt{[CLS]} token is passed through a linear projection
layer to produce a scalar score $s_i \in \mathbb{R}$ representing
boundary confidence.

The model is trained using binary cross-entropy loss on boundary
labels. No architectural modifications beyond the classification head
are introduced; the purpose of this implementation is to provide a
stable scoring signal rather than to optimize segmentation
performance.

\subsection{Training Procedure}

We train the scoring model in three stages, summarized in
Table~\ref{tab:training-protocol}:

\textbf{Stage 1: Synthetic Pretraining.}  We construct synthetic
splice examples by concatenating real dialogue segments and labeling
the splice point as a boundary. The synthetic splice corpus is
constructed from training splits of DailyDialog and MultiWOZ, ensuring no overlap with
validation or test data. We remove boundary-local artifacts (e.g.,
greetings) to prevent spurious cues, and we evaluate a negative
control set of single-segment examples to verify that the model does
not hallucinate boundaries. While some synthetic splice points may
join semantically related segments, this introduces conservative label
noise that discourages hallucinated boundaries rather than inducing
spurious segmentation.

\textbf{Stage 2: Supervised Fine-Tuning on Benchmarks.} 
We next perform supervised fine-tuning on benchmark
datasets. Table~\ref{tab:stage2} summarizes performance over five
epochs.

\textbf{Stage 3: Calibration and Final Evaluation.}
We apply temperature scaling to calibrate score magnitudes by
optimizing a single scalar temperature $T>0$ on a held-out validation
set (minimizing negative log-likelihood). This operation rescales the
logits used to produce boundary scores, improving score comparability
and numerical stability across runs. After calibration, the resulting
model is evaluated without further tuning on the test splits of all
datasets.

Importantly, temperature calibration does not explicitly enforce a target
boundary rate. Because it rescales scores, changing $T$ can affect
boundary density under a fixed threshold; however, explicit density
control requires selection parameters ($\tau$, $g$) or a controller.
Temperature calibration alone cannot enforce consistent segmentation
granularity across datasets.
Calibration therefore improves the interpretability of scores but does
not resolve granularity mismatch, which remains a decision-level issue
handled by explicit boundary selection.

\myparagraph{Reproducibility}
The reference implementation for this work is available in the
Episodic repository at \url{https://github.com/mhcoen/episodic}. The
\texttt{paper/} directory within this repository contains the LaTeX
source for the paper, experiment configuration files, and evaluation
scripts required to reproduce the reported results.

\myparagraph{Data availability}
All datasets used in this study are publicly available from their
original sources and are cited in the paper. Due to licensing
restrictions, we do not redistribute raw dialogue data. Instead, the
repository provides scripts to download, preprocess, and evaluate each
dataset in accordance with its original distribution terms.

\myparagraph{Protocol clarification}
Stage~2 supervised fine-tuning is performed exclusively on the
\emph{training splits} of SuperSeg and TIAGE; no other benchmark
datasets are used for supervised learning. Stage~3 temperature
calibration learns a single global temperature using a held-out
\emph{validation split} of SuperSeg only.

\begin{table}[t]
  \centering \small
  \begin{tabular}{llll}
    \toprule Stage & Purpose & Datasets & Split \\ \midrule Stage 1 &
    Synthetic pretraining & Synthetic splice corpus & Train / val
    \\ Stage 2 & Supervised fine-tuning & SuperSeg, TIAGE & Train (val for metrics)
    \\ Stage 3a & Temperature calibration & SuperSeg & Validation only
    \\ Stage 3b & Final evaluation & All 8 datasets & Test only
    \\ \bottomrule
  \end{tabular}
  \caption{Training, calibration, and evaluation protocol by stage.
    Stage~2 supervised fine-tuning uses only the training splits of
    SuperSeg and TIAGE to expose the scorer to both dense and sparse
    annotation regimes. Temperature calibration learns a single global
    temperature on a held-out validation split of SuperSeg. Stage~3
    evaluation is performed on test splits only, including datasets
    seen during supervised fine-tuning (SuperSeg, TIAGE), datasets
    used only for Stage~1 synthetic pretraining (DailyDialog, MultiWOZ),
    and datasets held out entirely (DialSeg711, Taskmaster, Topical-Chat,
    QMSum).}
  \label{tab:training-protocol}
\end{table}

\section{Evaluation Metrics}
\label{sec:evaluation-metrics}

We evaluate dialogue topic segmentation using metrics that separately
characterize boundary placement, boundary density, and gold-relative
segment alignment. This separation is necessary to distinguish
boundary localization errors from granularity mismatch and
segment-level structural effects that standard boundary-based scores
conflate.  Formal mathematical definitions are provided in
Appendices~A--C; the metrics are described here at an operational
level to support interpretation of the empirical results.

\myparagraph{$F_1$ over boundaries} We treat topic segmentation as
predicting a set of boundary indices $P_d \subseteq \{1,\dots,T_d-1\}$
and compare against gold boundaries $G_d \subseteq \{1,\dots,T_d-1\}$.
We compute precision $= |P_d \cap G_d|/|P_d|$, recall $=|P_d \cap
G_d|/|G_d|$, and $F_1$ as the usual harmonic mean.  For brevity, we
refer to this metric simply as $F_1$ throughout. Dataset-level F1 is
micro-averaged: $\mathrm{F1} = 2\sum_d|P_d \cap G_d| \;/\;
(\sum_d|P_d| + \sum_d|G_d|)$.

\myparagraph{Evaluation objectives}
Dialogue topic segmentation evaluation must distinguish between three
conceptually distinct properties: (i) boundary placement near
annotated topic transitions, (ii) alignment between predicted and
annotated boundary density (annotation regime), and (iii) internal
coherence of resulting segments with respect to the gold
segmentation. Standard boundary-based metrics collapse these
properties into a single score.  The evaluation framework used here
reports them separately.

\subsection{Window-tolerant boundary accuracy (W-F1)}

\myparagraph{What W-F1 measures}
W-F1 measures boundary placement accuracy under a fixed tolerance
window. A predicted boundary is counted as correct if it lies within a
small window around a gold boundary, rather than requiring an exact
index match. This makes W-F1 robust to small positional shifts that
are often operationally irrelevant. The variant used here adopts a
window-coverage convention, under which multiple predicted boundaries
may fall within the tolerance window of a single gold boundary and be
counted as correct. Formally, W-F1 is computed from dialogue-level precision and recall
defined over boundary sets $P_d$ and $G_d$ using a tolerance window
$\pm w$ (see Appendix~A for the full definition). In all experiments here, we use $w=1$.

\myparagraph{Interpretation and rationale}
This window-coverage design is intentionally duplicate-tolerant: multiple
predictions near the same gold boundary are not penalized as false positives.
Precision under W-F1 should therefore be interpreted as the fraction of
predicted boundaries that fall near annotated topic transitions, rather than
as a one-to-one measure of boundary set agreement. We adopt this convention
to decouple boundary localization from boundary count, allowing W-F1 to
serve as a diagnostic of whether boundaries are placed near annotated
transitions. For this reason, W-F1 must be interpreted in conjunction
with separate measures that account for boundary density and segment
structure.

\myparagraph{Relation to one-to-one tolerant matching (not used)}
A common alternative is \emph{one-to-one} tolerant matching, where each
gold boundary may be matched to at most one prediction within $\pm w$.
Under this convention, additional predicted boundaries near the same
gold transition are counted as false positives, making precision
sensitive to boundary count as well as boundary location. Because this
couples localization with boundary density, one-to-one tolerant
matching is not well suited to isolating boundary placement behavior
under varying segmentation granularities.

\myparagraph{Robustness to tolerant matching variants}
Although our primary analysis uses many-to-one window-coverage
matching, we additionally evaluate a standard one-to-one tolerant
matching variant (maximum bipartite matching within the same tolerance
window); see Appendix~G to verify that the
density-driven effects reported in this paper are not an artifact of
the matching convention. Across both datasets, the resulting
density--quality curves remain qualitatively unchanged
(Figure~G.1).

\myparagraph{Scope of the definition}
This W-F1 definition should not be assumed equivalent to other
window-tolerant F1 variants in prior work. A complete,
reimplementable definition—including matching rules, empty-set
handling, and aggregation—is given in Appendix~A.

\subsection{Boundary density (BOR)}
Boundary Over-segmentation Ratio (BOR) measures the number of
predicted boundaries relative to the number of gold
boundaries. Although termed the boundary oversegmentation ratio, BOR
is used here as a general boundary density ratio: values greater than
1 indicate oversegmentation relative to gold, while values less than 1
indicate undersegmentation. Formally, BOR is defined as the ratio of
total predicted boundaries to total gold boundaries across all
dialogues: $\mathrm{BOR} = \sum_{d} |P_d| \;/\; \sum_{d} |G_d|$. This
micro-aggregation reflects overall boundary density rather than being
skewed by high-variance ratios in short dialogues.

Boundary-based accuracy metrics alone cannot distinguish boundary
placement quality from boundary density alignment. A system can achieve
high W-F1 simply by matching the annotation boundary count, even with
non-semantic boundary placement. BOR exposes this effect directly by
making boundary density explicit.

\subsection{Segment alignment: purity and coverage}
\label{sec:segment-alignment}

\myparagraph{Definition and rationale}
Purity and coverage measure how well predicted segments align with the
gold segmentation, independent of exact boundary placement. Purity asks
whether each predicted segment draws most of its turns from a single
gold segment (limited cross-gold mixing), while coverage asks whether
each gold segment is largely captured by a single predicted segment
(limited fragmentation). Together, these measures distinguish between
adding boundaries that primarily refine gold segments and boundaries
that cut across or fragment them irregularly.

\myparagraph{Aggregation}
We report turn-weighted (micro) purity and coverage, computed per
dialogue and then aggregated across dialogues. Unweighted
segment-level (macro) variants are not monotone under refinement and
are therefore not used. Formal definitions and aggregation rules are
given in Appendix~B.

\subsection{Failure-mode interpretation}
The three reported diagnostics---W-F1, BOR, and purity/coverage---must be
interpreted jointly. W-F1 captures boundary misalignment under window-based
matching. BOR ($|P|/|G|$) exposes systematic over- or under-segmentation. Purity
and coverage are \emph{gold-relative segment alignment} diagnostics: they measure
how predicted and gold segments relate structurally (limited cross-gold mixing
versus fragmentation/subdivision), and should not be interpreted as independent
semantic coherence estimates.

Table~\ref{tab:failure-taxonomy} summarizes the characteristic regimes that arise
under different combinations of boundary accuracy, boundary density, and
purity/coverage.

\begin{table}[t]
  \centering
  \begin{tabular}{lccc}
    \toprule
    F1/W-F1 & BOR & Purity/Coverage & Interpretation \\
    \midrule
    High & $\approx 1$ & High (both) & Calibrated (aligned granularity) \\
    High & $\gg 1$ & High purity & Oversegmentation (fine-grained, gold-consistent) \\
    Low  & $\gg 1$ & High purity & Granularity mismatch (not necessarily noisy) \\
    Low  & $< 1$ & High coverage & Undersegmentation (coarser-than-gold) \\
    Low  & $\approx 1$ & Low (both) & Detection failure (misplaced/noisy) \\
    \bottomrule
  \end{tabular}
  \caption{\textbf{Failure modes of boundary-based evaluation.}
    Distinct segmentation behaviors can yield similar boundary accuracy scores
    unless boundary density (BOR) and purity/coverage are reported.}
  \label{tab:failure-taxonomy}
\end{table}

The baseline experiments in the following section illustrate several
of these regimes.

\subsection{Baselines}
\label{sec:baselines}
We evaluate simple non-semantic baselines to illustrate how boundary
density alone can dominate boundary-based evaluation. These baselines
provide reference points for interpreting density-driven effects
independent of semantic boundary placement.

We consider three classes of baselines: (i) a no-boundary baseline,
(ii) periodic segmentation baselines (\emph{Periodic Every-$N$}), which place
boundaries at fixed intervals with $N$ chosen per-dataset, and (iii) oracle-density
random baselines. We additionally evaluate an \emph{Oracle-Periodic} baseline that
matches the gold boundary count per dialogue exactly (used in Tables~\ref{tab:external_methods}
and~\ref{tab:leaderboard-deltas}).

Purity and coverage are reported here as diagnostic statistics rather than as
standalone evaluation objectives.
In particular, increasing boundary density tends to increase purity and decrease
coverage regardless of semantic validity.

\begin{table}[t]
  \centering
  \begin{tabular}{l l c c c c}
    \toprule
    Baseline & Dataset & W-F1 & \textbf{BOR} & Purity & Coverage \\
    \midrule 
    No-boundary & DialSeg711 & 0.000 & \textbf{0.00} & 0.438 & 1.000 \\
    & SuperSeg & 0.113 & \textbf{0.00} & 0.568 & 1.000 \\ 
    & TIAGE & 0.140 & \textbf{0.00} & 0.541 & 1.000 \\ 
    \midrule 
    Periodic Every-$N$ & DialSeg711 ($N=4$) & 0.639 & \textbf{0.97} & 0.812 & 0.823 \\
    & SuperSeg ($N=2$) & 0.623 & \textbf{1.10} & 0.830 & 0.775 \\
    & TIAGE ($N=3$) & 0.610 & \textbf{0.97} & 0.849 & 0.774 \\ 
    \midrule 
    Oracle-density Random & DialSeg711 & 0.528 & \textbf{1.00} & 0.839 & 0.833 \\
    (BOR=1) & SuperSeg & 0.678 & \textbf{1.00} & 0.883 & 0.884 \\ 
    & TIAGE & 0.664 & \textbf{1.00} & 0.873 & 0.869 \\ 
    \bottomrule
  \end{tabular}
  \caption{\textbf{Baseline diagnostic statistics relative to gold annotations.}
    Periodic (Every-$N$) places boundaries at fixed intervals, with $N$ chosen 
    per-dataset to approximate gold density. Oracle-density baselines match gold 
    boundary counts per dialogue by construction. Where non-zero, W-F1 for no-boundary
    reflects dialogues lacking gold boundaries (W-F1=1 under empty-set handling;
    see Appendix~A).}
  \label{tab:baselines}
\end{table}

These results instantiate several regimes from
Table~\ref{tab:failure-taxonomy}, including density-matched baselines
achieving competitive W-F1 despite lacking semantic grounding.

\section{Results}
\label{sec:results}
This section evaluates the proposed separation between boundary
scoring and boundary selection under differing annotation
granularity. The results below instantiate multiple failure modes
summarized in Table~\ref{tab:failure-taxonomy}, including results for
previously published methods re-evaluated under the same framework.

\subsection{Granularity as a Controlled Variable}
\label{sec:granularity}
To isolate the effect of segmentation granularity from boundary ranking quality,
we evaluate each method across a range of boundary densities by sweeping the
boundary selection threshold while holding the scoring model fixed.
This produces \emph{density--quality curves} that explicitly expose how boundary-based
metrics vary as a function of output density rather than at a single operating point.

\begin{figure}[!ht]
  {\small DialSeg711 and SuperSeg: Density--Quality Regime Plots with 95\% CIs}\\[1pt]
  \centering
  \includegraphics[width=\linewidth]{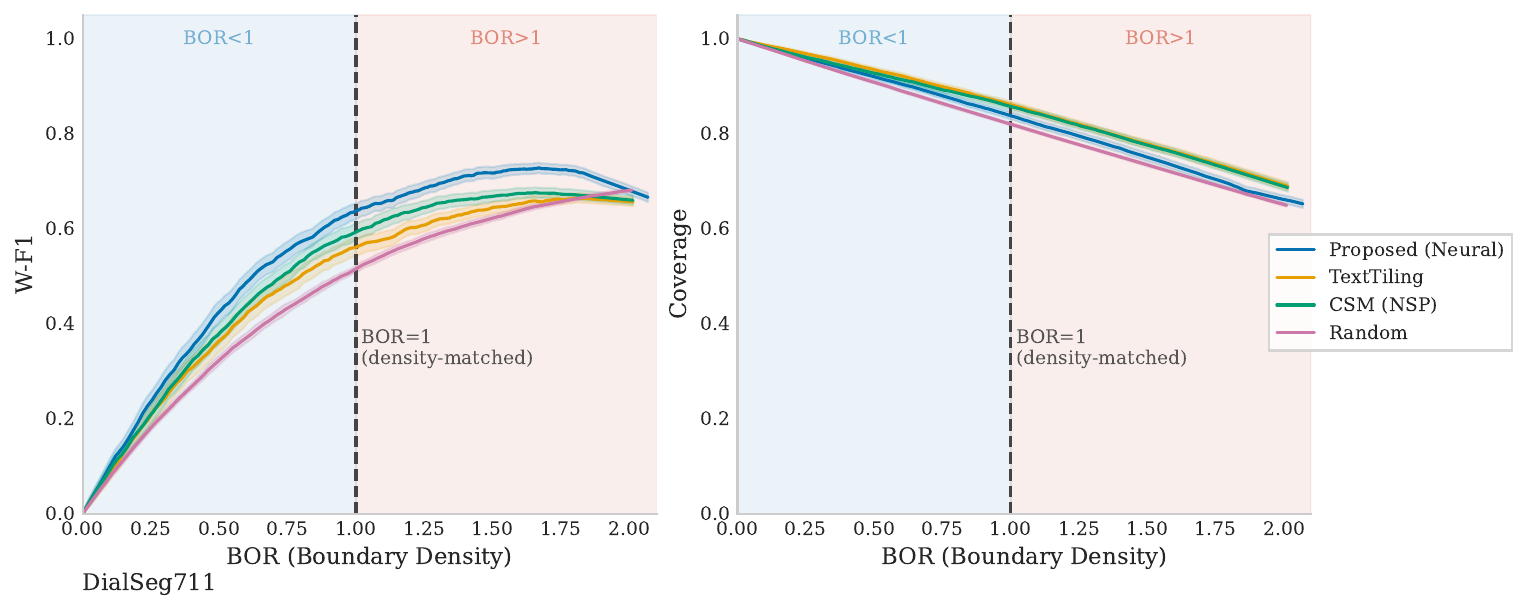}\\
  \vspace{-12pt}
  \noindent\makebox[\linewidth][c]{%
    \leaders\hbox{\rule{6pt}{0.3pt}\hspace{4pt}}\hskip0.9\linewidth
  }
  \vspace{4pt}
  \includegraphics[width=\linewidth]{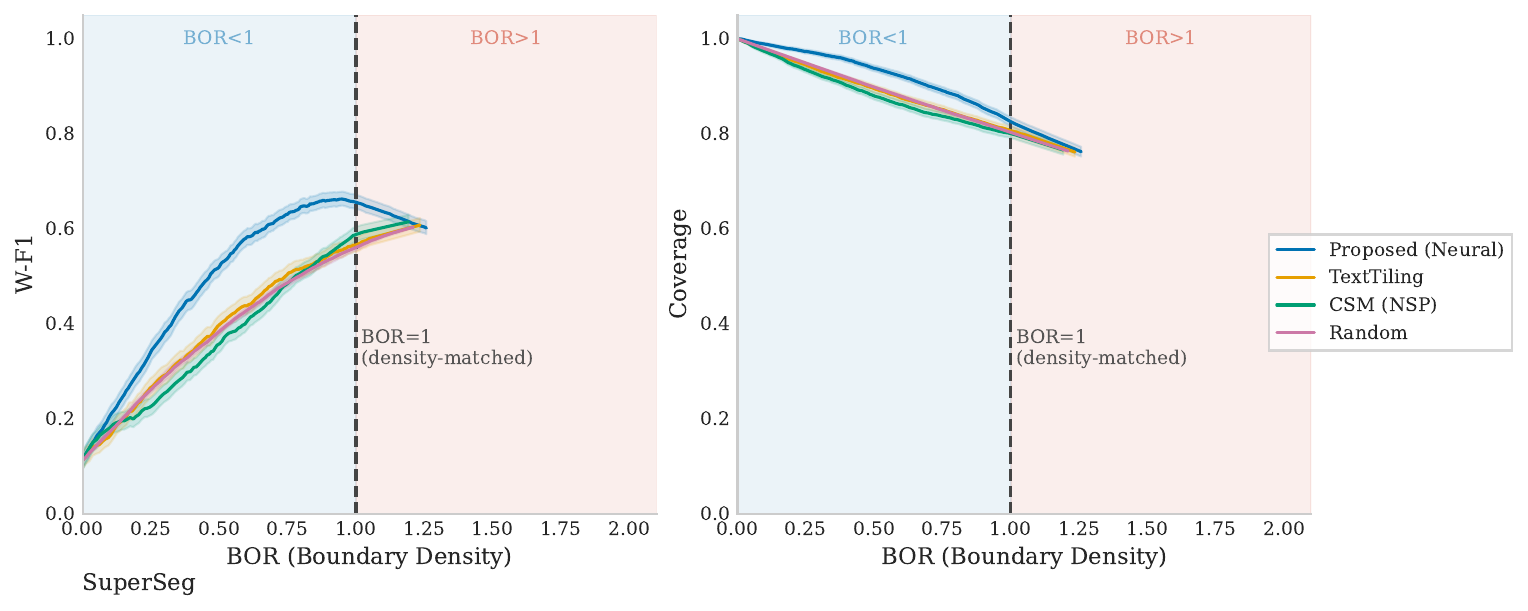}

  \caption{\textbf{Density--quality curves for DialSeg711 and
      SuperSeg.} Each curve varies the boundary-selection threshold
    for a fixed scoring model. Horizontal movement (changing boundary
    density) produces substantially larger changes in W-F1 and
    Coverage than vertical separation between methods, including for
    the proposed neural scorer. SuperSeg exhibits a comparatively
    narrow optimum near BOR~$\approx 1$, while DialSeg711 tolerates a
    broader range of boundary densities, indicating dataset-dependent
    coupling between boundary-based metrics and annotation
    granularity. Shaded regions surrounding each curve indicate 95\%
    dialogue-level bootstrap confidence intervals; curves are shown
    over the full realized BOR range for each dataset under the same
    threshold sweep. Results reflect the Episodic neural boundary
    scorer evaluated under static selection and are diagnostic of
    evaluation behavior rather than full deployed system performance.}
  \label{fig:density-quality}
\end{figure}

Figure~\ref{fig:density-quality} presents these curves for DialSeg711
and SuperSeg, illustrating two qualitatively distinct annotation
regimes.

\myparagraph{Interpreting density--quality curves}
Each density--quality curve in Figure~\ref{fig:density-quality}
traces a \emph{single}
segmentation method under different boundary-selection thresholds.
Moving horizontally along a curve changes only the \emph{number} of
output boundaries, as measured by BOR, while holding the underlying
scorer fixed. Vertical separation between curves at a matched BOR
therefore reflects differences in boundary \emph{ranking} quality,
whereas horizontal movement along a curve reflects changes in
segmentation granularity induced by selection. If boundary-based
metrics primarily measured boundary detection quality, methods would
be cleanly separated vertically at matched BOR; instead, the dominant
variation is typically along the BOR axis, indicating strong coupling
between W-F1/Coverage and boundary density.

\myparagraph{Dataset-dependent density sensitivity}
Figure~\ref{fig:density-quality} highlights a sharp contrast between
SuperSeg and DialSeg711. On SuperSeg, W-F1 is maximized near
BOR~$\approx 1$ and degrades rapidly under under- or
over-segmentation, indicating that evaluation strongly enforces the
annotation boundary density. On DialSeg711, by contrast, W-F1
exhibits a broad plateau across BOR, and oversegmentation is
comparatively weakly penalized. This contrast implies that identical
boundary-based metrics can encode substantially different effective
granularity assumptions across datasets.

Across both datasets, the proposed neural scorer is consistently
higher than baselines at matched BOR, particularly in the conservative
regime (BOR~$<1$), indicating improved boundary ranking.

\subsection{External Methods Under Granularity-Aware Evaluation}
To assess whether the granularity effects identified in this work are
specific to our scoring model or reflect broader evaluation behavior,
we re-evaluated a representative subset of published dialogue
segmentation methods under the proposed granularity-aware
framework. The selected methods include simple baselines (random and
periodic segmentation), a classical unsupervised approach
(TextTiling), and a coherence scoring model (CSM;
\citealt{xing2021improving}), which uses BERT's Next Sentence Prediction (NSP)
task to score coherence between adjacent utterances.  All methods were evaluated using the
same boundary canonicalization and metrics as in previous sections.

\myparagraph{External-method re-evaluations: scope and controls}
All external-method results use the same boundary representation and tolerance
window $w$, and all metrics (F1, W-F1, BOR, purity, coverage) are computed by a
single implementation under a shared canonicalization
(Appendix~D).
We therefore interpret
Table~\ref{tab:external_methods} as follows: under our canonicalization and
operating-point protocol, apparent single-number method differences largely
coincide with BOR shifts.

These re-evaluations are not intended as reproductions of published leaderboard
numbers. Reported scores can depend on underspecified or inconsistent details
across papers, including preprocessing (tokenization/turn handling and boundary
canonicalization), dataset splits, and operating-point selection; code and tuning
procedures are also not always available. A reproduction-oriented setup would
therefore entangle boundary density with these factors.

Instead, we evaluate all methods within a single canonicalized
pipeline with controlled density sweeps, so that BOR is an explicit,
comparable axis across methods. This design isolates the effect of
boundary density by holding canonicalization and selection behavior
fixed across methods. The audit protocol and canonicalization details
are provided in Appendix~H.

All single-point results reported below correspond to individual operating points
along density--quality curves (see Figure~\ref{fig:density-quality} for DialSeg711; TIAGE curves follow the same pattern).

\begin{table}[t]
  \centering
  \begin{tabular}{l l c c c c c}
    \toprule Method & Dataset & W-F1 & \textbf{BOR} & F1 & Purity &
    Coverage \\ \midrule TextTiling & DialSeg711 & 0.613 &
    \textbf{2.69} & 0.488 & 0.988 & 0.638 \\ CSM & DialSeg711 & 0.393
    & \textbf{0.56} & 0.353 & 0.732 & 0.931 \\ Random & DialSeg711 &
    0.267 & \textbf{0.55} & 0.098 & 0.711 & 0.890 \\ Oracle-Periodic &
    DialSeg711 & 0.481 & \textbf{1.00} & 0.211 & 0.806 & 0.809
    \\ \midrule TextTiling & TIAGE & 0.628 & \textbf{1.84} & 0.434 &
    0.933 & 0.717 \\ CSM & TIAGE & 0.409 & \textbf{0.58} & 0.306 &
    0.760 & 0.909 \\ Random & TIAGE & 0.256 & \textbf{0.33} & 0.121 &
    0.697 & 0.921 \\ Oracle-Periodic & TIAGE & 0.633 & \textbf{1.00} &
    0.276 & 0.824 & 0.826 \\ \bottomrule
  \end{tabular}
  \caption{\textbf{External dialogue segmentation methods re-evaluated
      under the proposed granularity-aware framework.} Oracle-Periodic
    places boundaries at evenly spaced intervals with count matched to
    gold per dialogue (BOR=1.00 by construction). Boundary density (BOR)
    varies substantially across methods, producing distinct density
    regimes despite comparable boundary accuracy scores. All
    threshold-based methods (TextTiling, CSM) evaluated at fixed
    operating points; no per-dataset tuning.}
  \label{tab:external_methods}
\end{table}

Several patterns in Table~\ref{tab:external_methods} align with the
failure modes summarized in Table~\ref{tab:failure-taxonomy}. Simple
periodic segmentation achieves competitive boundary accuracy when its
output density aligns with the gold annotation (BOR~$\approx 1$),
despite lacking semantic grounding. Classical methods such as
TextTiling exhibit recall-dominated behavior with elevated BOR,
yielding high purity alongside aggressive oversegmentation. CSM
operates conservatively on both datasets (BOR~$\approx 0.56$--$0.58$),
producing high coverage but lower purity---a precision-dominated
pattern characteristic of undersegmentation.

For clarity, and following the failure taxonomy in
Table~\ref{tab:failure-taxonomy}, we categorize the results in
Table~\ref{tab:external_methods} as conservative (BOR~$<1$), balanced
(BOR~$\approx 1$), or aggressive (BOR~$\gg 1$), as summarized in
Table~\ref{tab:regime_summary}.

\begin{table}[t]
  \centering
  \begin{tabular}{l l l}
    \toprule
    Method & Dataset & Density Regime \\
    \midrule
    TextTiling & DialSeg711 & Aggressive (BOR $\gg 1$) \\
    CSM        & DialSeg711 & Conservative (BOR $< 1$) \\
    Random     & DialSeg711 & Conservative (BOR $< 1$) \\
    Oracle-Periodic & DialSeg711 & Balanced (BOR $\approx 1$) \\
    \midrule
    TextTiling & TIAGE      & Aggressive (BOR $\gg 1$) \\
    CSM        & TIAGE      & Conservative (BOR $< 1$) \\
    Random     & TIAGE      & Conservative (BOR $< 1$) \\
    Oracle-Periodic & TIAGE & Balanced (BOR $\approx 1$) \\
    \bottomrule
  \end{tabular}
  \caption{Density regimes under the proposed failure taxonomy, derived from BOR thresholds.
    Regime classifications based on fixed operating points (no per-dataset tuning).}
  \label{tab:regime_summary}
\end{table}

\myparagraph{Threshold tuning and density side effects}
The results in Tables~\ref{tab:external_methods} and~\ref{tab:regime_summary}
use fixed internal thresholds for TextTiling and CSM, isolating intrinsic
method behavior from threshold effects. Standard practice, however, tunes
thresholds on a development set to maximize F1. We find that such tuning
can produce dramatically different operating points across datasets. For
example, when CSM's threshold is tuned per-dataset, the method shifts from
conservative (BOR~$=0.64$) on DialSeg711 to aggressive (BOR~$=3.97$) on
TIAGE---a sixfold increase in boundary density. With a fixed threshold,
CSM operates conservatively on both datasets (BOR~$\approx 0.56$--$0.58$),
revealing that the regime variation is a tuning artifact rather than inherent
method behavior. F1-based tuning selects boundary density as a side effect;
the resulting operating point can vary substantially across datasets
depending on annotation density, which is precisely the confound that
BOR-aware evaluation is designed to expose.

\myparagraph{Targeted audit of runnable SuperDialseg leaderboard comparisons}
We audit three runnable comparisons drawn from the SuperDialseg benchmark
leaderboard by re-evaluating the corresponding publicly available
implementations under our canonicalized pipeline and metrics (same tolerance
window $w$ as above). Although the comparisons originate from the SuperDialseg
leaderboard, the audited methods are evaluated here on DialSeg711 and TIAGE (both
included in SuperDialseg's evaluation suite). In all three cases, the reported
F1/W-F1 deltas coincide with higher boundary density (positive
$\Delta\mathrm{BOR}$), consistent with the density-regime effect analyzed above.

These three comparisons exhaust the set of SuperDialseg leaderboard entries that
we can reproduce from publicly available, deterministic implementations without
retraining and without proprietary or deprecated APIs. We therefore restrict the
audit to methods with runnable code under these constraints. In particular, the
InstructGPT baseline reported in SuperDialseg uses
\texttt{text-davinci-003}, which has since been deprecated and is no longer
available, so we do not include it here.

\begin{table}[t]
  \centering
  \small
  \setlength{\tabcolsep}{6pt}
  \begin{tabular}{llcccc}
    \toprule
    Comparison & Dataset &
    $\Delta$F1 &
    \parbox[c]{1.9cm}{\centering $\Delta$W-F1\\[0.5ex][95\% CI]} &
    \parbox[c]{1.9cm}{\centering $\Delta$BOR\\[0.5ex][95\% CI]} &
    Density Shift \\
    \midrule
    TT vs.\ CSM & DialSeg711 &
    +0.135 &
    \parbox[c]{1.9cm}{\centering +0.220\\[0.5ex][0.202, 0.237]} &
    \parbox[c]{1.9cm}{\centering +2.13\\[0.5ex][2.07, 2.18]} &
    \parbox[t]{2.6cm}{\raggedright Conservative $\rightarrow$\\Aggressive} \\
    \addlinespace[4pt]
    TT vs.\ CSM & TIAGE &
    +0.128 &
    \parbox[c]{1.9cm}{\centering +0.219\\[0.5ex][0.155, 0.283]} &
    \parbox[c]{1.9cm}{\centering +1.26\\[0.5ex][1.11, 1.44]} &
    \parbox[t]{2.6cm}{\raggedright Conservative $\rightarrow$\\Aggressive} \\
    \addlinespace[4pt]
    TT vs.\ Oracle-P & DialSeg711 &
    +0.277 &
    \parbox[c]{1.9cm}{\centering +0.132\\[0.5ex][0.114, 0.150]} &
    \parbox[c]{1.9cm}{\centering +1.69\\[0.5ex][1.64, 1.74]} &
    \parbox[t]{2.6cm}{\raggedright Balanced $\rightarrow$\\Aggressive} \\
    \bottomrule
  \end{tabular}
  \caption{\textbf{Pairwise audit of runnable SuperDialseg leaderboard comparisons with
      95\% dialogue-level bootstrap confidence intervals.} TT denotes TextTiling and
    Oracle-P is the Oracle-Periodic baseline (boundary count matched to gold per dialogue). Positive $\Delta$ indicates
    higher values for the first-listed method. Arrows denote relative changes in
    boundary density from the second-listed method to the first-listed method.
    Regime labels follow Table~\ref{tab:regime_summary}.
    All methods evaluated at fixed operating points.}
  \label{tab:leaderboard-deltas}
\end{table}

\myparagraph{Concrete example of a likely confounded comparison}
Table~\ref{tab:leaderboard-deltas} provides a direct illustration of how
published boundary-metric ``improvements'' can be driven by density
regime shifts rather than improved boundary placement. For example, on
DialSeg711, the comparison \emph{TextTiling vs.\ CSM} shows a substantial
increase in W-F1 ($\Delta$W-F1 $= +0.220$) accompanied by a large
increase in boundary density ($\Delta$BOR $= +2.13$), moving from a
conservative regime (BOR~$<1$) to an aggressive regime (BOR~$\gg 1$).
Under the failure taxonomy in Table~\ref{tab:failure-taxonomy}, this
pattern is consistent with recall-dominated oversegmentation: higher
tolerant boundary scores arise from emitting many more boundaries, not
necessarily from more accurate boundary localization at a matched
granularity.

\subsection{Stage 1: Synthetic Pretraining on Splice Boundaries}
This stage isolates whether the scoring model can learn a meaningful
boundary signal in a controlled setting, independent of annotation
noise. Performance here reflects the model's ability to distinguish
genuine topic transitions from non-boundaries before exposure to real
dialogue benchmarks. Table~\ref{tab:stage1} summarizes boundary-level
behavior during this phase.

\begin{table}[t]
  \centering
  \begin{tabular}{ccccccl}
    \toprule
    Epoch & F1 & W-F1 & BOR & Mean $|$score$|$ & Neg.\ Ctrl.\\
    \midrule
    1 & 0.265 & 0.420 & 0.31 & 0.007 & 0.000 \\
    2 & 0.453 & 0.688 & 0.67 & 0.009 & 0.032 \\
    3 & 0.489 & 0.733 & 0.87 & 0.010 & —  \\
    \bottomrule
  \end{tabular}
  \caption{\textbf{Stage 1 synthetic pretraining on splice boundaries.}
    Mean $|$score$|$ denotes the mean boundary score magnitude
    $\mathbb{E}[|s_i|]$ on the validation set and serves as a diagnostic
    for score collapse. Negative control (Neg.\ Ctrl.) reports the
    predicted boundary rate on single-segment dialogues, which should
    be near zero; it was not rerun for epoch~3, as the model had already
    passed this sanity check at epoch~2.
  }
  \label{tab:stage1}
\end{table}

\subsection{Stage 2: Supervised Fine-Tuning on Benchmarks}
This stage performs supervised fine-tuning to align the scoring model
with human annotation conventions on in-distribution datasets. The key
question is whether boundary density stabilizes without degrading
boundary ranking quality. Table~\ref{tab:stage2} reports performance
over five fine-tuning epochs. Supervised fine-tuning is therefore
in-distribution only for SuperSeg and TIAGE. Results on these datasets
in Stage~3 reflect performance after exposure to their annotation
regimes, whereas results on all other datasets reflect cross-dataset
generalization under annotation-granularity shift.

\begin{table}[t]
  \centering
  \begin{tabular}{c c c c c c}
    \toprule
    Epoch & \multicolumn{3}{c}{Overall (both datasets)} & SuperSeg & TIAGE \\
    \cmidrule(lr){2-4} \cmidrule(lr){5-5} \cmidrule(lr){6-6}
     & F1 & W-F1 & \textbf{BOR} & F1 / BOR & F1 / BOR \\
    \midrule
    1 & 0.778 & 0.833 & \textbf{1.11} & 0.788 / 1.13 & 0.197 / 0.24 \\
    2 & 0.819 & 0.850 & \textbf{1.23} & 0.830 / 1.25 & 0.352 / 0.86 \\
    3 & 0.840 & 0.865 & \textbf{1.13} & 0.852 / 1.14 & 0.337 / 0.82 \\
    4 & 0.848 & 0.875 & \textbf{1.08} & 0.862 / 1.08 & 0.322 / 0.78 \\
    5 & 0.843 & 0.872 & \textbf{1.04} & 0.857 / 1.05 & 0.311 / 0.84 \\
    \bottomrule
  \end{tabular}
  \caption{\textbf{Stage 2 supervised fine-tuning on benchmark datasets.}
    Metrics are computed on the held-out validation split (not used for gradient updates).
    ``Overall'' F1 and BOR are micro-averaged across both datasets;
    W-F1 is macro-averaged over dialogues (Appendix~A).
    Boundary density converges toward annotated density (BOR $\approx 1$)
    on datasets seen during training, while W-F1 remains high.}
  \label{tab:stage2}
\end{table}

\subsection{Stage 3: Final Test with Calibration}
\label{sec:stage3}

\myparagraph{Final test protocol}
Stage~3 evaluation is performed on the \emph{test splits} of all eight
datasets. This includes datasets seen during Stage~2 training
(SuperSeg, TIAGE) as well as datasets not used for supervised training
or calibration (DialSeg711, MultiWOZ, DailyDialog, Taskmaster,
Topical-Chat, QMSum). No dataset-specific thresholds, temperatures, or
selection parameters are tuned at this stage.

This stage evaluates generalization under distribution shift, where
annotation granularity differs from the training regime. Here, the
focus is not peak F1 but the interaction between boundary ranking,
boundary density, and calibration.

Final test results are shown in Table~\ref{tab:stage3}, which reports
evaluation metrics for a representative subset of three datasets
spanning sparse, medium, and dense annotation regimes; results on the
remaining datasets follow the same qualitative pattern.

\begin{table}[t]
  \centering
  \begin{tabular}{lcccc}
    \toprule
    Dataset & W-F1 & \textbf{BOR} & F1 & Pred./Gold Boundaries \\
    \midrule
    DialSeg711 & 0.767 & \textbf{2.53} & 0.434 & 5167/2042 \\
    SuperSeg   & 0.584 & \textbf{0.81} & 0.609 & 2376/2923 \\
    TIAGE      & 0.512 & \textbf{0.76} & 0.368 & 157/207 \\
    \bottomrule
  \end{tabular}
  \caption{\textbf{Stage 3 final test results.}
    Boundary density (BOR) is reported alongside boundary accuracy to
    distinguish granularity mismatch from detection failure. (F1 is micro-averaged; W-F1 is macro-averaged. See Appendix~A.)}
  \label{tab:stage3}
\end{table}

\noindent\textbf{Calibration.}
We learned a single global temperature of $T = 0.976$ for this run. In
practice, temperature scaling modestly improves score calibration and
threshold stability, but it does not eliminate cross-dataset threshold
non-transfer: differences in annotation granularity continue to
dominate the relationship between score thresholds and boundary
density.

Complete Stage~3 calibrated results for all eight datasets are reported
in Appendix~E.

\section{Gold-Relative Segment Alignment Under Granularity Mismatch}
These diagnostics are essential for interpreting boundary-based metrics under
granularity mismatch. In particular, they distinguish between two qualitatively
different outcomes that exact-match metrics conflate: (1) segmentations that
introduce additional boundaries that predominantly \emph{subdivide gold
segments} (high purity), and (2) segmentations that place boundaries in ways that
\emph{mix} turns from multiple gold segments or yield unstable
purity/coverage under matched density.

As shown in Table~\ref{tab:purity_coverage}, the proposed model maintains
high purity across datasets even when its boundary density exceeds that
of the gold annotation. Relative to random baselines with matched
density (BOR = 1), this indicates that the additional boundaries are
model-induced and predominantly subdivide gold segments, rather than
reflecting density-matched random placement. In contrast, baseline
methods that match gold density achieve competitive boundary scores but
exhibit lower or unstable purity/coverage, showing that density
alignment alone does not guarantee usable segmentation.

Taken together, these results suggest that the scorer shares the
annotators' inductive bias: it recovers many annotated boundaries and,
when producing finer segmentations ($\mathrm{BOR}>1$), predominantly
subdivides within gold segments rather than cutting across them. We
treat this as alignment with the gold topic partition, not as an
independent guarantee of semantic coherence.

\begin{table}[t]
  \centering
  \begin{tabular}{l c c c c c}
    \toprule
    Dataset & W-F1 & \textbf{BOR} & Purity & Coverage & F1 \\
    \midrule
    DialSeg711 & 0.767 & \textbf{2.53} & 0.962 & 0.651 & 0.434 \\
    SuperSeg   & 0.584 & \textbf{0.81} & 0.847 & 0.915 & 0.609 \\
    TIAGE      & 0.512 & \textbf{0.76} & 0.785 & 0.896 & 0.368 \\
    \bottomrule
  \end{tabular}
  \caption{\textbf{Gold-relative segment alignment diagnostics for the proposed model.}
    High purity with elevated BOR indicates fine-grained subdivision that
    largely preserves the gold partition; high coverage with reduced BOR
    indicates coarser-than-gold segmentation. Purity and coverage are
    gold-relative diagnostics and should not be interpreted as independent
    semantic coherence estimates. Both patterns reflect granularity
    mismatch rather than boundary detection failure.}
  \label{tab:purity_coverage}
\end{table}

\subsection{Worked Example}
\label{sec:worked_example}
\begin{figure}[t]
  \centering
  \includegraphics[width=\linewidth]{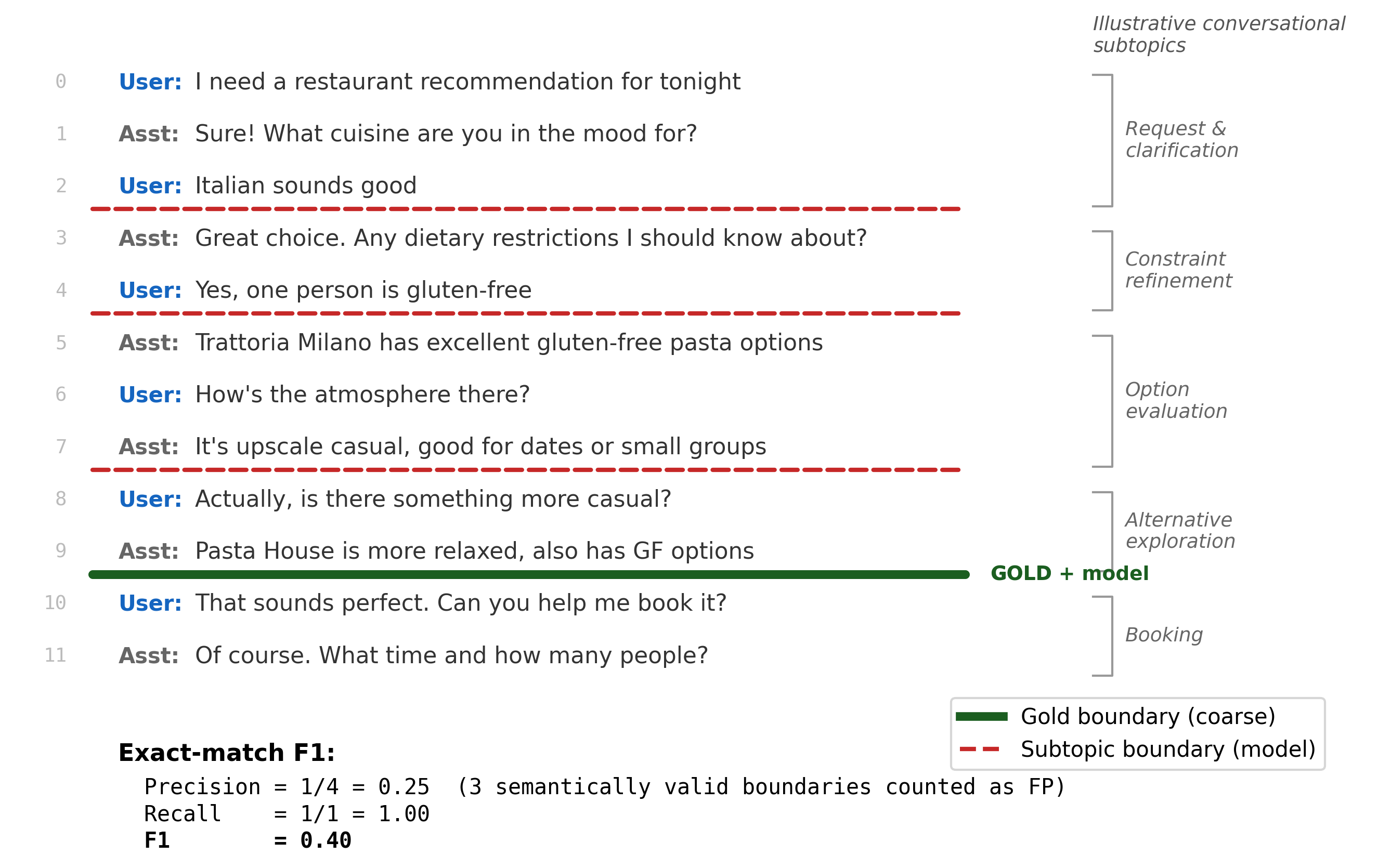}
  \caption{
    \textbf{Granularity mismatch in dialogue topic segmentation.}
    Gold annotations mark a single coarse topic boundary (browsing
    $\rightarrow$ booking), while the model predicts multiple additional
    boundaries associated with finer-grained discourse transitions.
    Under exact-match evaluation, three additional predicted boundaries
    are counted as false positives, yielding $F_1{=}0.40$ despite
    correct identification of the coarse transition. The discrepancy
    reflects a mismatch in boundary density and segmentation
    granularity rather than gross boundary detection error.}
  \label{fig:worked_example}
\end{figure}

Figure~\ref{fig:worked_example} isolates the granularity-mismatch
regime: the model predicts additional boundaries that are not annotated
in the (coarser) gold segmentation. Under strict precision/recall,
these extra predicted boundaries are treated as false positives,
reducing $F_1$ even in settings where purity/coverage indicate limited
cross-gold mixing (cf.\ Table~\ref{tab:purity_coverage}).

This example makes the failure mode concrete. The additional predicted
boundaries reflect finer-grained discourse transitions according to the
model, but they occur within a gold span and are therefore penalized by
exact-match evaluation. As a result, $F_1$ can drop substantially
despite correct identification of the major topic transition. The
dominant discrepancy is thus one of boundary density and segmentation
granularity rather than gross boundary misplacement, motivating
evaluation that distinguishes granularity mismatch from detection
failure.

\section{Discussion}

The results demonstrate that boundary-based accuracy metrics conflate
segmentation quality with boundary density alignment when annotation
granularity varies. In particular, the apparent competitiveness of
oracle-density baselines in Table~\ref{tab:baselines} should not be
interpreted as modeling capability, but as evidence that boundary
accuracy metrics reward density alignment rather than transferable
boundary detection. Random and periodic strategies achieve competitive
W-F1 and purity scores when their output boundary density is matched to
the gold annotation, despite lacking any semantic grounding. In
contrast, the proposed model yields substantially higher segment purity
on datasets with coarse annotations, at the cost of elevated BOR,
indicating limited cross-topic mixing relative to gold spans rather
than detection failure.

The strength of this density--metric coupling varies across datasets
depending on annotation consistency. When gold annotations are
relatively uniform across dialogues (e.g., DialSeg711), granularity
errors concentrate in a small subset of atypical dialogues. In this
setting, emitting additional boundaries is often beneficial, and W-F1
continues to improve under moderate oversegmentation.

When gold annotations are highly variable (e.g., SuperSeg),
granularity mismatch is distributed more evenly across dialogues.
Here, performance is tightly coupled to the annotated density, and
W-F1 is locally flat near the gold boundary rate.

These regime-specific behaviors explain the distinct shapes of the
density--quality curves and show that boundary-based metrics entangle
density alignment with segmentation quality in different ways across
datasets.

Figure~\ref{fig:density-quality} reinforces this point by showing that
the variation in W-F1 induced by sweeping the boundary selection
threshold for a fixed scoring model (moving along a curve) is
substantially larger than the variation observed between segmentation
methods at a matched boundary density.
\emph{This behavior is contrary to the standard evaluation assumption
that boundary-based metrics primarily reflect boundary placement
quality once density is controlled.}
We intentionally do not prescribe a universal target BOR, as any such
prescription would reintroduce a single ``correct'' granularity that
this work argues does not exist.

Importantly, this evaluation failure is not model-specific under these
benchmarks and evaluation protocols: any method whose output boundary
density is tuned to match the gold annotation can achieve competitive
F1 and W-F1 scores, as demonstrated by the non-semantic baselines in
Table~\ref{tab:baselines}. We do not attempt a comprehensive
retrospective survey of published claims across all leaderboards and
datasets; instead, we provide targeted, runnable audits
(Table~\ref{tab:leaderboard-deltas}) that concretely demonstrate how
apparent boundary-metric gains can arise from substantial shifts in
boundary density regimes. Adding more segmentation models does not
resolve this problem; it only illustrates it further.

Prior work and common usage patterns of classical change-detection and
drift-based strategies are consistent with this conclusion. Under
standard boundary-based evaluation protocols, such strategies are
typically recall-driven: they increase boundary density to improve
tolerant boundary metrics such as W-F1, frequently yielding boundary
oversegmentation ratios well above~1.0. The proposed evaluation
objective makes this trade-off explicit: methods with higher W-F1 but
elevated BOR are identified as oversegmenting, while more conservative
scorers exhibit lower boundary accuracy but better density control.

At a conceptual level, three qualitatively distinct boundary-density
regimes can be distinguished:
\begin{itemize}
\item \textbf{Conservative strategies}, which undersegment and produce
  $\text{BOR} < 1$, yielding lower recall but stable coverage.
\item \textbf{Balanced strategies}, which achieve $\text{BOR} \approx
  1$ with high purity and coverage, yielding segmentation aligned with
  the annotation granularity.
\item \textbf{Aggressive strategies}, which substantially increase
  boundary density ($\text{BOR} \gg 1$) to maximize recall and tolerant
  boundary scores, resulting in recall-dominated detection rather than
  calibrated segmentation.
\end{itemize}

In practice, selection thresholds should therefore be chosen based on
intended downstream use rather than optimized solely to match a
particular annotation density. Boundary density metrics such as BOR
provide a simple diagnostic for tuning selection behavior without
retraining the scoring model.

\myparagraph{Practical guidance for choosing boundary density}
The appropriate boundary selection behavior depends on how segmentation
outputs are used downstream. Based on the empirical results in this
study, we offer the following high-level guidance:

\begin{itemize}
\item \textbf{Summarization and context truncation.}
  When topic boundaries trigger summarization or removal of prior
  turns, missing a major boundary can be more harmful than introducing
  extra minor boundaries. Conservative selection is therefore
  preferable: a lower boundary density ($\text{BOR} < 1$) reduces the
  risk of prematurely discarding relevant context, even if some
  distinct topics are merged.

\item \textbf{Retrieval and navigation.}
  For applications that rely on topic boundaries to index or retrieve
  relevant conversation segments, finer-grained segmentation can be
  beneficial. In such settings, semantic coherence is an engineering
  objective of the topic segmentation algorithm itself rather than a
  quantity directly evaluated here. When boundary density increases
  ($\text{BOR} > 1$) but purity remains high, the additional boundaries
  predominantly reflect within-gold subdivision rather than cross-gold
  cuts, yielding finer-grained retrieval units consistent with the gold
  topic partition.

\item \textbf{Exploratory analysis and annotation.}
  When segmentation is used for qualitative analysis, annotation
  support, or exploratory inspection, fine-grained boundaries are
  often desirable. In such settings, boundary density may substantially
  exceed that of existing annotations, and evaluation should emphasize
  purity/coverage over exact boundary alignment.
\end{itemize}

These guidelines reinforce the central claim of this paper: boundary
density is not an intrinsic property of a conversation, but a design
choice that should be tuned to application requirements rather than
optimized to match a given annotation scheme. As
Section~\ref{sec:results} demonstrates, F1-based threshold tuning
implicitly searches for the density that matches gold annotations
rather than improving boundary localization. This dependence on
selection behavior motivates treating boundary selection as a
distinct, explicit step.

\subsection{Boundary Selection as a Separate Step}
\label{sec:boundary-selection}
\begin{figure}[t]
  \centering
  \includegraphics[width=\textwidth]{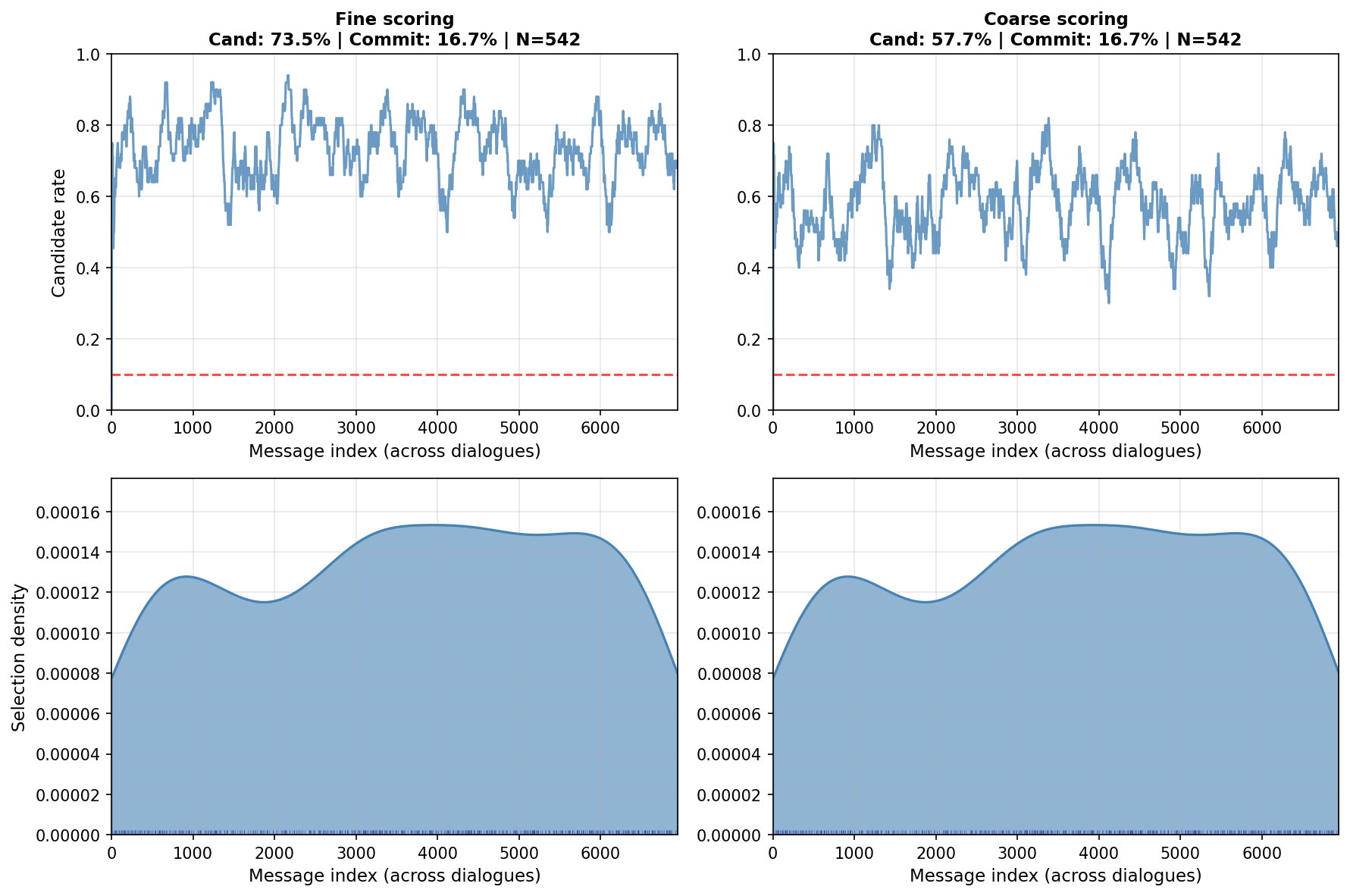}
  \caption{\textbf{Adaptive boundary selection enforces invariant
      boundary distributions across scoring granularities
      (DialSeg711).}
    \emph{Top row} shows the raw candidate boundary rate produced by the
    same scorer under fine (left) and coarse (right) candidate
    thresholds, resulting in substantially different candidate
    frequencies (73.5\% vs.\ 57.7\%). The dashed line indicates the
    target selection rate specified to the controller. \emph{Bottom
    row} shows the resulting distribution of selected boundaries over
    message index after adaptive thresholding. Despite large
    differences in candidate generation, adaptive selection converges
    to nearly identical boundary distributions and total selection
    counts (16.7\% selection rate, $N{=}542$ in both cases). This
    invariance is not the result of score rescaling or post-hoc
    normalization: the underlying scoring function is unchanged, and
    the effect emerges from online evidence accumulation and adaptive
    thresholding, reflecting successful decoupling of boundary
    selection from boundary scoring. The adaptive selection procedure
    illustrated here is formally specified in
    Appendix~F.}
  \label{fig:adaptive_commitment}
\end{figure}

\noindent\textbf{Scope.} All offline evaluations
(Sections~\ref{sec:evaluation-metrics}--\ref{sec:results}) use the
static threshold\(+\)gap rule over one-shot scores
(\S\ref{sec:selection-rule}). The adaptive evidence-accumulating
controller (Appendix~F) is used only for the control
experiment in this subsection (Figure~\ref{fig:adaptive_commitment}).

Figure~\ref{fig:adaptive_commitment} illustrates the consequence of
separating scoring from selection: different scoring granularities
produce substantially different candidate boundary rates, yet an
explicit selection rule enforces stable output boundary distributions
that are invariant to the scorer's granularity. Crucially, this
invariance does not arise from score rescaling or post-hoc
normalization. The scoring function is unchanged; identical output
distributions emerge solely from the selection mechanism's evidence
accumulation and adaptive thresholding behavior. This behavior
illustrates that separating scoring from selection enables boundary
density to be controlled independently of scoring granularity,
motivating its treatment as a design parameter rather than an
evaluation artifact. A fully specified and reimplementable version of
this adaptive controller is given in Appendix~F.

\subsection{Computational Cost}

Boundary selection is computationally negligible: it operates on a
vector of scores and applies thresholding and minimum spacing. Boundary
scoring dominates cost because it requires running the neural model
over candidate positions (window-vs-window inputs around each
position). The separation therefore has practical value in deployed
systems: selection can be retuned per dataset or application without
recomputing scores, while recomputing scores requires model inference.
Evidence accumulation within online boundary selection is likewise
lightweight relative to scoring, since it reuses the fixed-reference
scores and maintains only decision state.

\subsection{On Dataset Choice and External Validity}

Public dialogue datasets are necessarily imperfect proxies for real
human--AI interactions. While several conversational AI datasets have
been released, we are not aware of any publicly available corpus of
long, open-ended human--AI conversations with turn-level topic
boundary annotations suitable for evaluating segmentation granularity.
Our claims therefore concern evaluation behavior under heterogeneous
annotation schemes rather than direct modeling of user behavior. The
consistency of failure modes across diverse datasets indicates that
these effects arise from evaluation practice rather than
dataset-specific artifacts.

\myparagraph{Human--LLM chat data and annotation availability}
\label{sec:chatdata}
Few large-scale corpora of human--LLM interaction are publicly
available, and none currently include turn-level topic boundary
annotations suitable for segmentation evaluation. WildChat
\cite{zhao2024wildchat}, for example, contains over one million
authentic user--assistant exchanges but is released without discourse-
or topic-level segmentation labels. Nevertheless, such data exhibit
strong initiative asymmetries: users introduce goals and direct topic
shifts, while assistant turns are reactive responses constrained by the
preceding query. This asymmetry is well established in studies of
institutional and task-oriented interaction \cite{drew1992talk} and
has been shown to shape discourse segmentation under mixed-initiative
conditions \cite{walker1990mixed}. In deployed AI assistant
systems, these asymmetries are typically addressed through
systems-level design choices for intent tracking and context
management \cite{chopra2024exploring}, which operate downstream of the
boundary detection problem studied here.

Given the absence of annotated human--LLM dialogue data, the present
study does not model speaker-specific segmentation policies. Instead,
it examines how boundary-based evaluation metrics behave under varying
annotation granularities using existing human--human and task-oriented
dialogue corpora with explicit topic labels. Alternative segmentation
unit choices (e.g., restricting candidates to user turns only) reflect
engineering considerations for deployed systems and primarily rescale
the set of candidate boundary positions; they do not fundamentally
alter the evaluation phenomena analyzed here, which concern the
interaction between boundary density, placement quality, and segment
structure relative to gold annotations. Empirical validation on
annotated human--LLM data remains an important direction for future
work.

\section{Limitations and Future Work}
This work contributes (i) an evaluation objective that makes
boundary density and segment coherence explicit and (ii) a topic
segmentation algorithm that separates boundary scoring from boundary
selection. We do not explore alternative neural backbones or include a
direct ablation comparing synthetic pretraining against training from
scratch. Preliminary experiments indicate that splice-boundary
pretraining primarily stabilizes early training dynamics rather than
improving final performance, but a systematic ablation remains an
important direction for follow-up work.

Purity and coverage are computed with respect to the same gold
annotations whose granularity we argue is often mismatched to model
behavior. Accordingly, these measures should not be interpreted as
independent estimates of semantic coherence. Rather, they function as
relative diagnostics: high purity rules out incoherent mixing of gold
topics, while elevated BOR distinguishes fine-grained subdivision from
boundary noise. Independent coherence measures (e.g., embedding-based
similarity or human judgments) would be valuable complements, but are
not required for the present goal of diagnosing evaluation failure
modes under mismatched annotation granularity.

\myparagraph{Interpretation constraint}
Purity and coverage should be interpreted only conditional on
non-trivial boundary detection performance (e.g., W-F1 significantly
greater than zero). When considered in isolation, both measures can be
arbitrarily high or low for segmentations with no semantic content,
including density-matched random baselines.

\myparagraph{Statistical variability}
We report dialogue-level bootstrap confidence intervals for selected
figures and tables (e.g., Figure~\ref{fig:density-quality} and
Table~\ref{tab:leaderboard-deltas}) to characterize
variability arising from dialogue sampling under fixed model
checkpoints and boundary-selection parameters. These intervals are
computed by resampling dialogues and do not reflect variability across
random seeds, training runs, or alternative model initializations.

We do not report multi-run or multi-seed confidence intervals or formal
significance tests across independent training runs. Our primary
claims concern qualitative failure modes and regime-level behavior
(e.g., recall-dominated oversegmentation versus calibrated
segmentation), which manifest as large and consistent shifts in
boundary density and coherence that substantially exceed the
within-sample bootstrap variability observed here. Multi-seed
experiments indicated that run-to-run variance was small relative to
these effects and did not alter the qualitative conclusions. A
systematic multi-run statistical analysis remains a direction for
future work.

Inter-annotator agreement (IAA) statistics could also provide
additional empirical support for the claim that topic boundaries are
granularity-dependent rather than objective. Several dialogue
segmentation datasets report IAA in their original publications (e.g.,
DialSeg711~\cite{xu2021topicaware},
SuperSeg~\cite{jiang2023superdialseg}), often showing only moderate
agreement even under controlled annotation protocols. However, these
statistics are reported using heterogeneous definitions and units
(e.g., boundary placement, segment overlap, or task-level agreement),
making direct comparison across datasets difficult. Because the
present work focuses on evaluation behavior under fixed annotation
schemes rather than on the reliability of those schemes themselves, we
do not reanalyze annotator agreement here. Integrating IAA into a
unified analysis of granularity effects remains an important direction
for future work.

Additional directions include evaluating published dialogue
segmentation methods under the proposed evaluation objective to
confirm that the observed granularity effects are method-agnostic, and
exploring alternative boundary selection rules that adapt boundary
density to application constraints rather than annotation conventions.
Further study is also needed to relate boundary density preferences to
specific downstream uses such as summarization, retrieval, or
conversational memory. Finally, future work could replace discrete
boundary representations with continuous or multi-scale topic
representations (e.g., topic clouds) that better capture overlapping,
gradual, or revisited conversational structure.

\section{Conclusion}
\label{sec:conclusion}
Dialogue topic segmentation does not admit a single ground truth.
Exact boundary matching conflates detection correctness with
segmentation granularity and can obscure coherent segmentation behavior
under heterogeneous annotation schemes.
By evaluating boundary density and purity/coverage alongside W-F1,
this work provides a principled basis for diagnosing evaluation failure
modes that boundary accuracy metrics alone cannot distinguish.

A central empirical finding of this work is that boundary-based metrics
can be dominated by boundary density alignment rather than boundary
placement quality.
For example, under F1-tuned operating points, CSM occupied markedly
different density regimes across datasets (conservative on DialSeg711,
aggressive on TIAGE); under fixed operating points, its boundary
density was stable. The observed regime shift was a tuning artifact,
not an intrinsic method--dataset interaction---illustrating how
F1-based optimization can actively induce the granularity confounds
that standard evaluation protocols fail to diagnose.

Boundary-based accuracy metrics should not be interpreted at a single
operating point without accompanying density and purity/coverage.
More broadly, segmentation granularity should be treated as a design
choice conditioned on downstream use, not as an implicit property of a
benchmark annotation scheme.

\FloatBarrier
\clearpage

\appendix
\section{Window-tolerant F1 (W-F1)}
\label{app:wf1}

This appendix provides the formal mathematical definition of the
window-tolerant F1 metric used throughout the paper. The main text
remains fully interpretable without consulting this appendix.

Let a dialogue $d$ consist of $T_d$ turns, with gold boundary indices
$G_d \subset \{1,\dots,T_d-1\}$ and predicted boundary indices
$P_d \subset \{1,\dots,T_d-1\}$. Let $w \ge 0$ denote the tolerance window.

\myparagraph{Window-coverage matching}
Define the number of window-matched predicted boundaries
\[
\mathrm{TP}_d
= \left|\left\{
p \in P_d \;:\; \exists g \in G_d \text{ such that } |p-g| \le w
\right\}\right|,
\]
and the number of recovered gold boundaries
\[
\mathrm{MG}_d
= \left|\left\{
g \in G_d \;:\; \exists p \in P_d \text{ such that } |p-g| \le w
\right\}\right|.
\]
Under this window-coverage convention, multiple predicted boundaries may be
matched to the same gold boundary, while each gold boundary contributes at most
once to recall.

\myparagraph{Dialogue-level precision and recall}
Precision and recall are defined per dialogue with explicit empty-set handling:
\[
\mathrm{Prec}_d =
\begin{cases}
\mathrm{TP}_d / |P_d|, & |P_d| > 0, \\
1, & |P_d| = 0 \wedge |G_d| = 0, \\
0, & |P_d| = 0 \wedge |G_d| > 0,
\end{cases}
\]
\[
\mathrm{Rec}_d =
\begin{cases}
\mathrm{MG}_d / |G_d|, & |G_d| > 0, \\
1, & |G_d| = 0 \wedge |P_d| = 0, \\
0, & |G_d| = 0 \wedge |P_d| > 0.
\end{cases}
\]

\myparagraph{Dialogue-level W-F1}
The window-tolerant F1 score for dialogue $d$ is
\[
\mathrm{W\text{-}F1}_d =
\begin{cases}
\displaystyle
\frac{2\,\mathrm{Prec}_d\,\mathrm{Rec}_d}
     {\mathrm{Prec}_d + \mathrm{Rec}_d},
& \mathrm{Prec}_d + \mathrm{Rec}_d > 0, \\
0, & \text{otherwise}.
\end{cases}
\]

\myparagraph{Dataset-level aggregation}
Dataset-level W-F1 is computed as a macro average over dialogues:
\[
\mathrm{W\text{-}F1}
= \frac{1}{|D|} \sum_{d \in D} \mathrm{W\text{-}F1}_d .
\]

\myparagraph{Scope note}
This definition specifies the W-F1 variant used throughout this paper and should
not be assumed equivalent to other window-tolerant F1 formulations that employ
different matching or aggregation rules.

\myparagraph{Alternative (not used): one-to-one window matching}
For completeness, we define a one-to-one variant that credits at most one prediction
per gold boundary. Let $H_d$ be a bipartite graph with left nodes $P_d$, right nodes $G_d$,
and an edge $(p,g)$ iff $|p-g|\le w$. Let $M_d$ be a maximum-cardinality matching in $H_d$,
and define $TP^{1{:}1}_d \coloneqq |M_d|$. Then
\[
\mathrm{Prec}^{1{:}1}_d = \frac{TP^{1{:}1}_d}{|P_d|},\qquad
\mathrm{Rec}^{1{:}1}_d = \frac{TP^{1{:}1}_d}{|G_d|},
\]
with the same empty-set conventions as above, and
$\mathrm{W\text{-}F1}^{1{:}1}_d$ as the harmonic mean of
$\mathrm{Prec}^{1{:}1}_d$ and $\mathrm{Rec}^{1{:}1}_d$.
All results in this paper use the window-coverage definition presented here
(i.e., $TP_d$ and $MG_d$ as defined above), not the one-to-one variant.

\section{Purity and Coverage}
\label{app:purity-coverage}
This appendix defines the purity and coverage diagnostics used to
characterize gold-relative segment alignment in
Section~\ref{sec:evaluation-metrics}.

Let a dialogue $d$ be segmented into predicted segments
$P^{\mathrm{seg}}_d = \{p_1,\dots,p_m\}$ and gold segments
$G^{\mathrm{seg}}_d = \{g_1,\dots,g_n\}$, where each segment is a contiguous span
of turns and the segmentations form partitions of $\{1,\dots,T_d\}$.

For a segment $s = [a,b]$, let $|s| = b-a+1$ denote its length in turns, and for
two segments $s$ and $t$ define their overlap
\[
|s \cap t| = \max\!\left(0,\; \min(b_s,b_t) - \max(a_s,a_t) + 1\right).
\]

\myparagraph{Segment-level purity and coverage}
For a predicted segment $p \in P^{\mathrm{seg}}_d$, define its purity as
\[
\mathrm{Purity}(p)
= \max_{g \in G^{\mathrm{seg}}_d} \frac{|p \cap g|}{|p|}.
\]
For a gold segment $g \in G^{\mathrm{seg}}_d$, define its coverage as
\[
\mathrm{Coverage}(g)
= \max_{p \in P^{\mathrm{seg}}_d} \frac{|p \cap g|}{|g|}.
\]

\myparagraph{Dialogue-level micro averages}
Turn-weighted (micro) purity and coverage for dialogue $d$ are
\[
\mathrm{Purity}_d
= \frac{1}{T_d} \sum_{p \in P^{\mathrm{seg}}_d}
\max_{g \in G^{\mathrm{seg}}_d} |p \cap g|,
\]
\[
\mathrm{Coverage}_d
= \frac{1}{T_d} \sum_{g \in G^{\mathrm{seg}}_d}
\max_{p \in P^{\mathrm{seg}}_d} |p \cap g|.
\]

\myparagraph{Dataset-level aggregation}
Dataset-level purity and coverage are computed as macro averages over dialogues:
\[
\mathrm{Purity}
= \frac{1}{|D|} \sum_{d \in D} \mathrm{Purity}_d,
\qquad
\mathrm{Coverage}
= \frac{1}{|D|} \sum_{d \in D} \mathrm{Coverage}_d.
\]

\section{Monotonicity Properties}
\label{app:monotonicity}
This appendix formalizes monotonicity properties referenced in
Section~\ref{sec:evaluation-metrics}.

\myparagraph{Lemma (Purity is monotone under refinement)}
Let $P'_d$ be a refinement of $P_d$, obtained by adding predicted boundaries
(i.e., splitting predicted segments). Then
\[
\mathrm{Purity}_d(P'_d, G_d) \ge \mathrm{Purity}_d(P_d, G_d).
\]

\myparagraph{Proof}
It suffices to consider splitting a single predicted segment
$p \in P_d$ into subsegments $p_1,\dots,p_k$.
Let $g^\ast \in G^{\mathrm{seg}}_d$ be a gold segment maximizing overlap with $p$.
Since the $p_i$ partition $p$,
\[
|p \cap g^\ast| = \sum_{i=1}^k |p_i \cap g^\ast|.
\]
For each $i$,
$\max_{g} |p_i \cap g| \ge |p_i \cap g^\ast|$.
Summing over $i$ yields a non-decreasing contribution to
$\mathrm{Purity}_d$, while all other segments are unchanged.
\hfill$\square$

\myparagraph{Remark (Coverage under refinement)}
Under the same refinement,
$\mathrm{Coverage}_d(P'_d, G_d) \le$ $\mathrm{Coverage}_d(P_d, G_d)$, since splitting
predicted segments cannot increase the maximum overlap with any fixed gold
segment.

\section{Boundary canonicalization}
\makeatletter
\def\@currentlabelname{Window-tolerant F1 (W-F1)}
\makeatother
\label{app:canonicalization}

All datasets are converted to a canonical message sequence $m_1,\dots,m_T$ per dialogue, where each $m_t$ corresponds
to one dataset-provided utterance after filtering (below). Candidate boundary indices are the between-message positions
$i\in\{1,\dots,T-1\}$.

\myparagraph{Filtering}                                                                 
We (i) retain all dialogue turns from both speakers (user and agent/assistant), removing only dataset-level metadata headers where present (e.g., meeting       
preambles in QMSum), (ii) preserve speaker role labels but do not include them in utterance text, and (iii) remove empty or whitespace-only utterances if any   
exist. After filtering, messages are reindexed so that all boundary indices refer to positions in the filtered sequence.                                        

\myparagraph{Gold boundary extraction}
For each dataset, we define the gold boundary set $G_d\subseteq\{1,\dots,T_d-1\}$ using the dataset's native segment
annotation as follows:
\begin{itemize}
\item \textbf{Datasets with segment IDs per utterance:} let $\ell_t$ be the segment ID for message $m_t$; then
  $G_d = \{\, i : \ell_i \ne \ell_{i+1}\,\}$.
\item \textbf{Datasets with explicit boundary markers:} let $b_t\in\{0,1\}$ indicate that $m_t$ starts a new segment;
  then $G_d = \{\, i : b_{i+1}=1\,\}$.
\item \textbf{Domain-/state-labeled datasets:} let $\texttt{label}_t$ denote the dataset-provided label for message $m_t$; then
  $G_d=\{\, i : \texttt{label}_i \ne \texttt{label}_{i+1}\,\}$.
\end{itemize}

\myparagraph{Dataset-specific notes}
Table~\ref{tab:canon} summarizes the canonicalization rules; complete
extraction implementations are in the accompanying code repository.

{\hyphenpenalty=10000
\exhyphenpenalty=10000

\begin{table}[t]
\centering
\small
\setlength{\tabcolsep}{6pt}
\begin{tabularx}{\linewidth}{
l
c
>{\raggedright\arraybackslash}X
>{\raggedright\arraybackslash}X
}
\toprule
Dataset &
Unit &
Annotation $\rightarrow G_d$ rule &
Filtering notes \\
\midrule
DialSeg711 & U &
Segment ID change between consecutive utterances &
All turns retained; indexed consecutively \\

SuperSeg & U &
Segment ID change between consecutive utterances &
All turns retained; indexed consecutively \\

TIAGE & U &
Task/topic segment change between consecutive utterances &
All turns retained; indexed consecutively \\

MultiWOZ & U &
Domain label change between consecutive utterances &
All turns retained; indexed consecutively \\

DailyDialog & U &
Topic category change between consecutive utterances &
All turns retained; indexed consecutively \\

Taskmaster & U &
Task or subtask label change between consecutive utterances &
All turns retained; indexed consecutively \\

Topical-Chat & U &
Conversation topic change between consecutive utterances &
All turns retained; indexed consecutively \\

QMSum & T &
Section/topic boundary between adjacent speaker turns &
Meeting metadata removed \\
\bottomrule
\end{tabularx}
\caption{Boundary canonicalization details required to reproduce gold
boundary sets $G_d$ from each dataset.
Unit codes: \textbf{U} = each message is a boundary candidate (typical for two-party dialogues where speakers alternate);
\textbf{T} = boundaries occur only at speaker changes, grouping consecutive sentences from the same speaker into one unit (used for multi-party meeting transcripts like QMSum).}
\label{tab:canon}
\end{table}
}

\section{Full Stage~3 Calibrated Results Across All Datasets}
\label{app:stage3}

All results reported in this appendix use the identical Stage~3 evaluation pipeline
and are computed exclusively on \emph{test splits}:
a single model checkpoint after Stage~2 training,
a single global temperature calibrated on the SuperSeg validation split,
fixed boundary canonicalization,
and fixed boundary-selection parameters.
No information from any test split is used during training or calibration.
The evaluation protocol follows the train/validation/test split specification
summarized in Table~\ref{tab:training-protocol}.

The main text (Section~\ref{sec:stage3}) reports Stage~3 calibrated results on three
representative datasets spanning sparse, medium, and dense annotation
regimes. To substantiate the cross-dataset claims made in Sections~4 and~7,
this appendix reports the same Stage~3 evaluation
metrics for all eight datasets used in this study: DialSeg711,
SuperSeg, TIAGE, MultiWOZ, DailyDialog, Taskmaster,
Topical-Chat, and QMSum.

All results in Table~\ref{tab:stage3-all8} were computed using the
identical Stage~3 evaluation pipeline as the main Stage~3 results
(Table~\ref{tab:stage3}): a single globally calibrated temperature
($T=0.976$), identical boundary canonicalization, fixed boundary
selection parameters, and the evaluation definitions in Section~\ref{sec:evaluation-metrics}.

We report window-tolerant boundary detection (W-F1 with a $\pm 1$
message tolerance), boundary density relative to gold annotations
(BOR), and segment coherence diagnostics (purity and coverage), along
with predicted and gold boundary counts. Strict exact-match boundary
F1 is omitted here because it is highly sensitive to aggregation
choice (e.g., micro vs.\ macro averaging) and boundary sparsity, and
it is not a primary diagnostic in the proposed evaluation objective.

\begin{table}[t]
  \centering
  \small
  \begin{tabular}{lcccccc}
    \toprule
    Dataset & W-F1 & BOR & Purity & Coverage & Pred./Gold \\
    \midrule
    DialSeg711    & 0.767 & 2.53  & 0.962 & 0.651 & 5167/2042 \\
    SuperSeg      & 0.584 & 0.81  & 0.847 & 0.915 & 2376/2923 \\
    TIAGE         & 0.512 & 0.76  & 0.785 & 0.896 & 157/207   \\
    DailyDialog   & 0.398 & 0.91  & 0.783 & 0.885 & 182/200   \\
    MultiWOZ      & 0.738 & 2.22  & 0.950 & 0.757 & 2475/1117 \\
    Taskmaster    & 0.288 & 5.23  & 0.955 & 0.565 & 1026/196  \\
    Topical-Chat  & 0.289 & 3.91  & 0.939 & 0.620 & 766/196   \\
    QMSum         & 0.154 & 17.56 & 0.979 & 0.459 & 3705/211  \\
    \bottomrule
  \end{tabular}
  \caption{\textbf{Stage~3 calibrated results for all eight datasets.}
    Metrics computed under the same calibrated Stage~3 pipeline used in Table~\ref{tab:stage3},
    with identical model checkpoint, temperature calibration ($T=0.976$), boundary canonicalization,
    and selection settings.
    This appendix reports a subset of diagnostics focused on boundary density and segment coherence;
    strict exact-match F1 is intentionally omitted because it is highly sensitive to aggregation and
    boundary sparsity and is not a primary diagnostic in the proposed evaluation objective.}
  \label{tab:stage3-all8}
\end{table}

\section{Adaptive Boundary Selection}
\label{app:adaptive}

This appendix provides a formal specification of the adaptive boundary
selection mechanism referenced in Section~\ref{sec:boundary-selection}.

\myparagraph{Definitions and target rate}
Candidate boundaries correspond to between-message indices
$i \in \{1,\ldots,T{-}1\}$, where $i$ denotes the position between messages
$(m_i, m_{i+1})$.

Let $C(t)$ denote the number of candidate positions processed up to time $t$, and
let $S(t)$ denote the number of boundaries committed up to time $t$. The adaptive
controller targets a boundary \emph{selection rate}
\[
\rho \;=\; \mathbb{E}\!\left[\frac{S(t)}{C(t)}\right],
\]
measured per processed candidate position (not per message or per dialogue).

\myparagraph{Evidence accumulation}
Each candidate position $i$ maintains an accumulated evidence value $e_i(t)$.
When a candidate $i$ first becomes available, a boundary score $s_i$ is computed
once using frozen reference windows anchored at that time. This score remains
fixed thereafter.

As additional post-boundary context becomes available, evidence is accumulated
according to
\[
e_i(t{+}1) \;=\; e_i(t) + s_i .
\]
This implements a time-to-threshold integrator: candidates with higher initial
scores $s_i$ reach the commitment threshold faster than candidates with lower
scores. Freezing the score at the time of first availability ensures that the
commitment decision reflects the strength of the boundary signal at the point of
topic departure, rather than being confounded by subsequent context that may
diverge further or return to the original topic. The adaptive threshold
$\tau(t)$ (below) provides the primary density control mechanism.

\myparagraph{Boundary commitment rule}
At time $t$, a candidate boundary at position $i$ is committed if and only if
\[
e_i(t) \ge \tau(t)
\quad\text{and}\quad
|i - b| \ge g \;\;\forall b \in \mathcal{B},
\]
where $\tau(t)$ is the current decision threshold, $g$ is a minimum spacing
constraint, and $\mathcal{B}$ is the set of previously committed boundaries.
Candidates violating the spacing constraint are suppressed.

\myparagraph{Adaptive threshold controller}
The threshold $\tau(t)$ is updated online to enforce the target selection rate
$\rho$. Let $\hat{\rho}(t)$ denote the empirical selection rate over the most
recent $W$ processed candidates (or all candidates if fewer than $W$ have been
seen). The controller update is
\[
\tau(t{+}1) \leftarrow \tau(t) + \eta\,(\hat{\rho}(t) - \rho),
\]
where $\eta > 0$ is a step size. Increasing $\tau$ reduces boundary commits, while
decreasing $\tau$ increases them.

\myparagraph{Pseudocode}
The following pseudocode provides a complete operational realization of the
adaptive boundary selection procedure described above.

\begin{algorithm}[t]
\caption{Adaptive boundary selection}
\label{alg:adaptive}

\KwIn{target rate $\rho$, spacing $g$, window size $W$, step size $\eta$,
initial threshold $\tau(0)$}
\KwOut{committed boundary set $\mathcal{B}$}

$\mathcal{B} \leftarrow \emptyset$\;
Initialize counters $C \leftarrow 0$, $S \leftarrow 0$\;

\For{each time step $t$ as messages arrive}{
  \For{each newly available candidate $i$}{
    compute frozen score $s_i$; set $e_i \leftarrow 0$\;
  }
  \For{each active candidate $i$}{
    $e_i \leftarrow e_i + s_i$\;
    $C \leftarrow C + 1$\;
    \If{$e_i \ge \tau(t)$ \textbf{and} $|i-b| \ge g$ for all $b \in \mathcal{B}$}{
      $\mathcal{B} \leftarrow \mathcal{B} \cup \{i\}$\;
      $S \leftarrow S + 1$\;
    }
  }
  compute $\hat{\rho}(t)$ over last $W$ candidates\;
  $\tau(t{+}1) \leftarrow \tau(t) + \eta(\hat{\rho}(t)-\rho)$\;
}
\end{algorithm}

\myparagraph{Remarks}
The adaptive controller operates on candidate positions rather than messages,
making the target rate invariant to dialogue length. Freezing the boundary score
avoids repeatedly re-scoring shifting contexts, and ensures that high-scoring
candidates are committed before low-scoring ones under time pressure imposed by
the adaptive threshold. When $\tau(t)$ is held fixed and accumulation is
disabled, the procedure reduces to the static selection rule used in the main
experiments.

\FloatBarrier

\section{One-to-one tolerant matching variant for W-F1}
\label{app:wf1:matching}
Let $P$ be predicted boundaries and $G$ gold boundaries, and let $w$ be the tolerance window.
Construct a bipartite graph with edges $(p,g)$ when $|p-g|\le w$.
Compute a maximum-cardinality matching $M \subseteq P \times G$ (ties broken arbitrarily or by minimum total distance).
Define $\mathrm{TP}=|M|$, $\mathrm{FP}=|P|-\mathrm{TP}$, and $\mathrm{FN}=|G|-\mathrm{TP}$, and compute precision/recall/F1 in the usual way.
This enforces the one-to-one constraint that each predicted (resp.\ gold) boundary can be credited at most once.

\begin{figure}[ht]
  \centering
  \includegraphics[width=\linewidth]{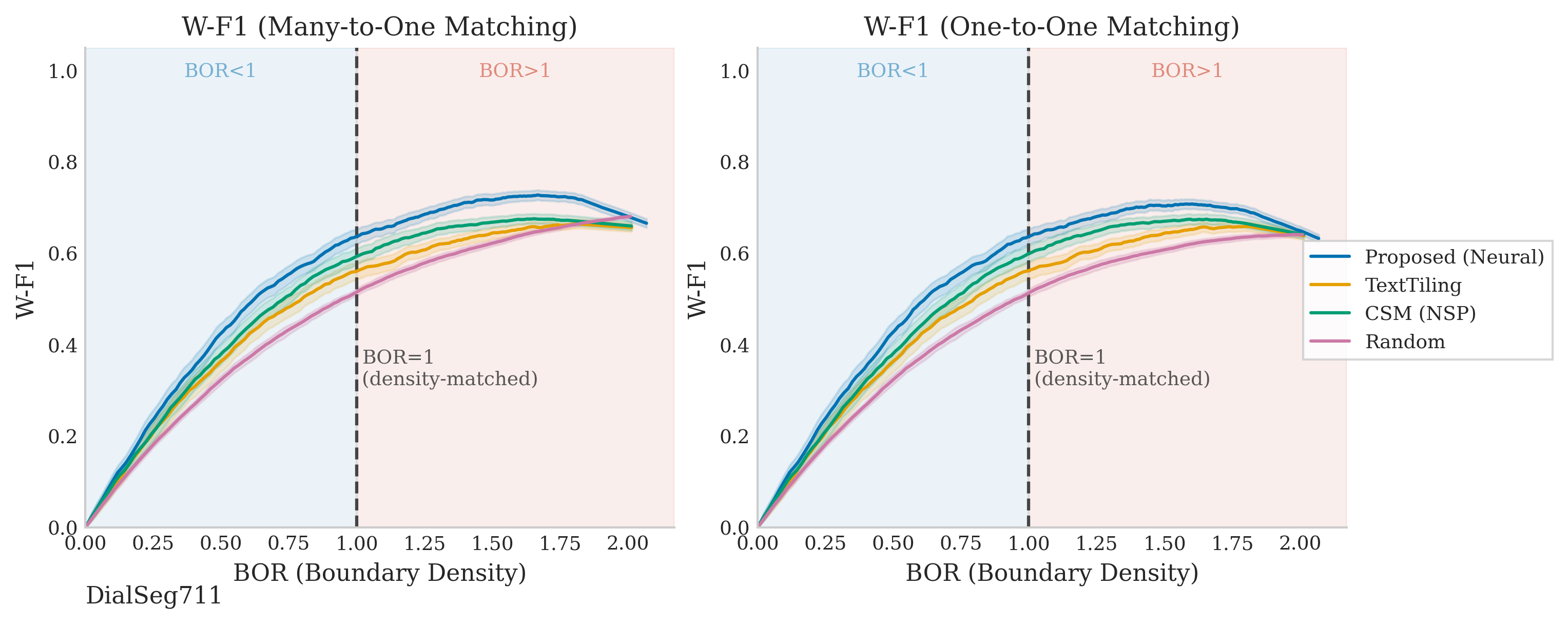}\\
\vspace{-12pt}
\noindent\makebox[\linewidth][c]{%
  \leaders\hbox{\rule{6pt}{0.3pt}\hspace{4pt}}\hskip0.9\linewidth
}
\vspace{4pt}
  \includegraphics[width=\linewidth]{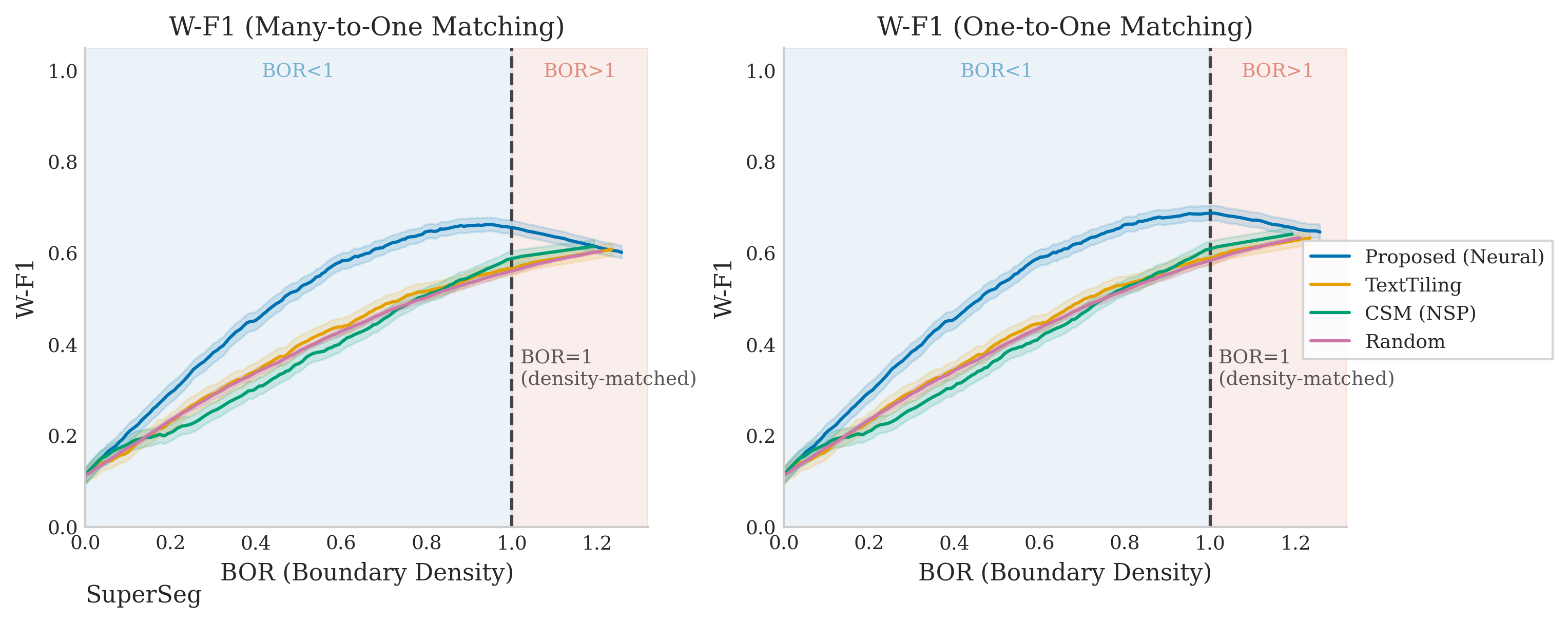}
  \caption{\textbf{Robustness of density--quality curves to tolerant matching.}
  Left: many-to-one (window-coverage) W-F1 used in the main paper.
  Right: one-to-one tolerant matching via maximum bipartite matching within the same window $w$.
  In both datasets, varying boundary density (BOR) induces the dominant changes in W-F1 under either matching, while method differences at matched BOR are secondary. Shaded regions surrounding each curve indicate 95\%
    dialogue-level bootstrap confidence intervals; curves are shown
    over the full realized BOR range for each dataset under the same
    threshold sweep. Results reflect the Episodic neural boundary
    scorer evaluated under static selection and are diagnostic of
    evaluation behavior rather than full deployed system performance.}
  \label{fig:matching_comparison}
\end{figure}

\section{Audit protocol and canonicalization details}
\label{app:audit}
\myparagraph{Controlled}
We fix: (i) boundary indexing and mapping to turn positions, (ii) the
tolerance window $w$, (iii) metric implementations, and (iv) the
operating-point protocol used to select or sweep thresholds to induce a
range of BOR values for each method.

\myparagraph{Not controlled}
We do not attempt to replicate each method's original preprocessing,
training regime, or originally reported threshold tuning; therefore, we
do not interpret discrepancies with published single-number reports as
errors, and we do not attribute differences uniquely to modeling choices
independent of preprocessing.
\FloatBarrier

\bibliography{references}

\end{document}